\documentclass{article}

\usepackage[final]{neurips_2021}

\usepackage[utf8]{inputenc} 
\usepackage[T1]{fontenc}    
\usepackage{url}            
\usepackage{booktabs}       
\usepackage{amsfonts}       
\usepackage{nicefrac}       
\usepackage{microtype}      
\usepackage{pifont}
\usepackage{graphicx}
\usepackage{amsmath, amsthm, amssymb, bm}
\usepackage{color, colortbl}
\usepackage[dvipsnames]{xcolor}
\usepackage{mathtools}
\usepackage{enumitem}
\usepackage{multirow}
\usepackage[ruled,vlined]{algorithm2e}
\usepackage{floatrow}
\usepackage{caption}
\usepackage{subcaption}
\usepackage{booktabs}
\usepackage{listings}
\usepackage{xcolor}
\usepackage{graphicx,wrapfig,lipsum}
\usepackage{tabularx}
\usepackage{hyperref}       

\newcolumntype{H}{>{\setbox0=\hbox\bgroup}c<{\egroup}@{}}

\newtheorem{theorem}{Theorem}[section]
\newtheorem{corollary}{Corollary}[theorem]
\newtheorem{lemma}[theorem]{Lemma}

\newcommand{\algo}{\texttt{POODLE}\xspace}
\newcommand{\longalgo}{\textbf{P}enalizing \textbf{O}ut-\textbf{O}f-\textbf{D}istribution samp\textbf{LE}s}
\newcommand{\etal}[1]{{{#1}} \textit{et al.}}
\newcommand{\ie}{\emph{i.e.,}~}
\newcommand{\eg}{\emph{e.g.,}~}
\newcommand{\wrt}{\emph{w.r.t.}~}
\newcommand{\myparagraph}[1]{\vspace{-6pt}\paragraph{#1}}

\usepackage[colorinlistoftodos, textwidth=30mm, shadow, textsize=small]{todonotes}
\definecolor{babyblue}{rgb}{0.54, 0.81, 0.94}
\definecolor{citrine}{rgb}{0.89, 0.82, 0.04}
\definecolor{misocolor}{rgb}{0.16,0.27,0.86}
\definecolor{jbcolor}{rgb}{0.9,0.4,0.2}
\definecolor{bernacolor}{rgb}{0.9608,0.4863,0.00}
\definecolor{carlcolor}{rgb}{0.0,0.9863,0.30}
\definecolor{grey}{rgb}{0.3, 0.3, 0.3}

\definecolor{graphicbackground}{rgb}{0.96,0.96,0.8}
\definecolor{rouge1}{RGB}{226,0,38}  
\definecolor{orange1}{RGB}{243,154,38}  
\definecolor{jaune}{RGB}{254,205,27}  
\definecolor{blanc}{RGB}{255,255,255} 
\definecolor{rouge2}{RGB}{230,68,57}  
\definecolor{orange2}{RGB}{236,117,40}  
\definecolor{taupe}{RGB}{134,113,127} 
\definecolor{gris}{RGB}{91,94,111} 
\definecolor{bleu1}{RGB}{38,109,131} 
\definecolor{bleu2}{RGB}{28,50,114} 
\definecolor{vert1}{RGB}{133,146,66} 
\definecolor{vert3}{RGB}{20,200,66} 
\definecolor{vert2}{RGB}{157,193,7} 
\definecolor{darkyellow}{RGB}{233,165,0}  
\definecolor{lightgray}{rgb}{0.9,0.9,0.9}
\definecolor{darkgray}{rgb}{0.6,0.6,0.6}
\definecolor{babyblue}{rgb}{0.54, 0.81, 0.94}
\definecolor{citrine}{rgb}{0.89, 0.82, 0.04}
\definecolor{misogreen}{rgb}{0.25,0.6,0.0}
\definecolor{PalePurp}{rgb}{0.66,0.57,0.66}
\definecolor{todocolor}{rgb}{0.66,0.99,0.99}
\definecolor{pearOne}{HTML}{2C3E50}
\definecolor{pearTwo}{HTML}{A9CF54}
\definecolor{pearTwoT}{HTML}{C2895B}
\definecolor{pearThree}{HTML}{E74C3C}
\colorlet{titleTh}{pearOne}
\colorlet{bull}{pearTwo}
\definecolor{pearcomp}{HTML}{B97E29}
\definecolor{pearFour}{HTML}{588F27}
\definecolor{pearFith}{HTML}{ECF0F1}
\definecolor{pearDark}{HTML}{2980B9}
\definecolor{pearDarker}{HTML}{1D2DEC}

\hypersetup{
	colorlinks,
	citecolor=pearDark,
	linkcolor=pearThree,
	breaklinks=true,
	urlcolor=pearDarker}

\definecolor{mydarkblue}{rgb}{0,0.08,1}
\definecolor{mydarkgreen}{rgb}{0.02,0.6,0.02}
\definecolor{mydarkred}{rgb}{0.8,0.02,0.02}
\definecolor{mydarkorange}{rgb}{0.40,0.2,0.02}
\definecolor{mypurple}{RGB}{111,0,255}
\definecolor{myred}{rgb}{1.0,0.0,0.0}
\definecolor{mygold}{rgb}{0.75,0.6,0.12}
\definecolor{myblue}{rgb}{0,0.2,0.8}
\definecolor{mydarkgray}{rgb}{0.66,0.66,0.66}

\title{POODLE: Improving Few-shot Learning via Penalizing Out-of-Distribution Samples}

\author{%
	Duong H. Le \thanks{First two authors contribute equally.} \\
	VinAI Research\\
	\texttt{v.duonglh5@vinai.io} \\
	\And
	Khoi D. Nguyen \footnotemark[1]\\
	VinAI Research\\
	\texttt{khoinguyenucd@gmail.com} \\
	\And
	Khoi Nguyen \\
	VinAI Research \\
	\texttt{ducminhkhoi@gmail.com}
	\AND
	Quoc-Huy Tran \\
	Retrocausal, Inc.\\
	\texttt{huy@retrocausal.ai} \\
	\And
	Rang Nguyen \\
	VinAI Research \\
	\texttt{rangnhm@gmail.com} \\
	\And 
	Binh-Son Hua \\
	VinAI Research \& \\ VinUniversity 
}

\newfloatcommand{capbtabbox}{table}[][\FBwidth]
\newcommand{\cmark}{\ding{51}}%
\newcommand{\xmark}{\ding{55}}%

\begin{document}
	
	\maketitle
    	\begin{abstract}
		In this work, we propose to leverage out-of-distribution samples, \ie unlabeled samples coming from outside target classes, for improving few-shot learning. Specifically, we exploit the easily available out-of-distribution samples (\eg from base classes) to drive the classifier to avoid irrelevant features by maximizing the distance from prototypes to out-of-distribution samples while minimizing that to in-distribution samples (\ie support, query data).
		Our approach is simple to implement, agnostic to feature extractors, lightweight without any additional cost for pre-training, and applicable to both inductive and transductive settings. Extensive experiments on various standard benchmarks demonstrate that the proposed method consistently improves the performance of pretrained networks with different architectures. Our code is available at \url{https://github.com/lehduong/poodle}.
	\end{abstract}	
	
\section{Introduction}


Learning with limited supervision is a key challenge to translate the research efforts of deep neural networks to real-world applications where large-scale annotated datasets are prohibitively costly to acquire. This issue has motivated the recent topic of few-shot learning (FSL), which aims to build a system that can quickly learn new tasks from a small number of labeled data. 		
	
A popular group of methods in FSL focus on strengthening the backbone network by various techniques, from increasing model capacity~\cite{chen2019closer, doersch2020crosstransformers}, self-supervised learning (SSL)~\cite{gidaris2019boosting, su2020does, zhang2021iept}, to knowledge distillation (KD)~\cite{tian2020rethinking}. 
With these techniques, few-shot methods are expected to learn better representations that are more robust and generalized. 
However, even if the network can discover visual features and semantic cues, few-shot learners have to deal with a key challenge - the \emph{ambiguity}: as we have only a small amount of support evidence, there are multiple plausible hypotheses at the inference stage. Existing works, therefore, rely on the developed \emph{inductive bias} of the network (during the pretraining stage), such as shape bias~\cite{ritter2017cognitive,feinman2018learning}, to reduce the hypothesis space. 

In this work, we view the classification problem as conditional reasoning, \ie \emph{``if X has P then X is Q''}. Human beings are good at learning such inferences, thus quickly grasping new concepts with minimal supervision.
More importantly, humans learn new concepts in context - where we have already had prior knowledge about other entities.
According to \emph{mental models} in cognitive science, when assessing the validity of an inference, one would retrieve counter-examples, \ie which do not lead to the conclusion despite satisfying the premise~\cite{de2005working, edgington1995conditionals, johnson2010mental, verschueren2005everyday}. Thus, if there exists at least one of such counter-examples, the inference is known to be erroneous.

Hence, we attempt to equip few-shot learning with the above ability so that it can eliminate incorrect hypotheses when learning novel tasks in a data-driven manner. Specifically, we leverage out-of-distribution data, \ie samples belonging to classes separated from novel tasks~\footnote{We use OOD samples and distractor samples interchangeably}, as counter-examples for preventing the learned prototypes from overfitting to their noisy features. To that end, when learning novel tasks, we adopt the \emph{large margin principle} in metric learning~\cite{weinberger2006distance} to encourage the learned prototypes to be close to support data while being distant from out-of-distribution samples. 

Our approach is complementary to existing works in FSL and could be combined to advance the state of the art. Moreover, our method is agnostic to the backbone network; thus, it does not have the need of a training phase to adopt as in SSL and KD, while incurring just a little overhead at inference (for fine-tuning prototypes with approximately 200 gradient updating steps).

In summary, our contributions are as follows:
\begin{itemize}[leftmargin=*]
    \vspace{-3pt}
	\item We propose a novel yet simple approach to learn  the inductive bias of deep neural networks for FSL by leveraging out-of-distribution data. We empirically show that out-of-distribution data only require weak labels (\ie in the form of whether a sample is in- or out-of-distribution) even in challenging problems such as cross-domain FSL.
	\item We introduce a new loss function to implement the above idea, which is applicable for both inductive and transductive inference. Our extensive experiments on different standard benchmarks show that the proposed approach consistently improves the performance of various network architectures.
	\item We validate the effectiveness of our method in various FSL settings, including cross-domain FSL. 
\end{itemize}

\begin{figure}[t]
	\centering
	\includegraphics[width=\linewidth, clip]{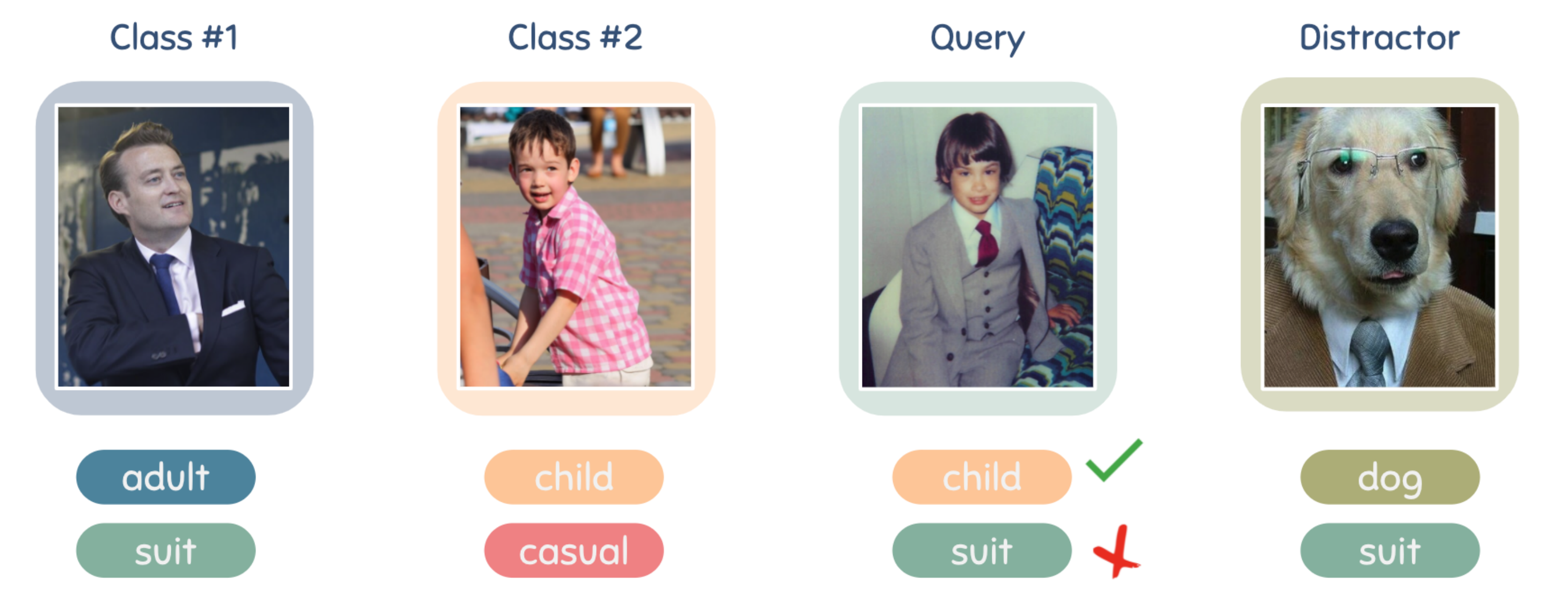}
	\caption{Illustration of advantages of counter-example data.
	The query image has features from both support classes (\ie a child and a suit), which makes classification ambiguous. The prediction result would depend on the inductive bias or prior knowledge of the network.  
	By using an out-of-distribution sample, it becomes clear that the query result should favor Class 2.}
	\label{fig:example}
	\vspace{-5pt}
\end{figure}

\section{Related work}
Over the past few years, considerable amount of research efforts~\cite{finn2017model, franceschi2018bilevel, jamal2019task, lee2018gradient, qiao2018few, matching_net, prototypical_nets, zhang2020deepemd} have been invested in FSL. They can roughly be classified into two main categories: \textit{optimization-based} and \textit{metric-based} approaches. Optimization-based methods~\cite{finn2017model, franceschi2018bilevel, jamal2019task, lee2018gradient, qiao2018few} seek a meta-learner that can quickly adjust the parameters of another learner to a new task, given only a few support images. In particular, \cite{li2017meta, finn2017model} propose learning to initialize a classifier whose parameters can be obtained with a small number of gradient updates on the novel classes. Metric-based approaches~\cite{matching_net,prototypical_nets,zhang2020deepemd} learn a task-agnostic embedding space for measuring the similarity between images. For example, Matching Networks~\cite{matching_net} utilizes a weighted nearest neighbor classifier, while Prototypical Networks \cite{prototypical_nets} uses the mean features of support images as the prototype for each class. Recently, DeepEMD \cite{zhang2020deepemd} adopts the Earth Mover's distance to compute the distance between image patches as their distance. 

	
To further boost the performance, recent works have incorporated additional techniques such as \textit{self-supervised learning} and \textit{knowledge distillation} during training, and \textit{transductive inference} during testing, which we will review in the following:

\textbf{Self-supervised learning.} The goal of self-supervised learning is to learn representations from unlabeled data. For FSL, this could be achieved by combining the supervised main task with a self-supervised pretext task which includes predicting image rotations~\cite{gidaris2018unsupervised}, predicting relative patch locations~\cite{doersch2017multi}, or solving jigsaw puzzles~\cite{noroozi2016unsupervised}. A few works incorporating self-supervised learning for FSL have been developed recently, e.g., Gidaris et al.~\cite{gidaris2019boosting} suggest pre-training the embedding network by using a combination of supervised and self-supervised loss, \ie predicting image rotations. 

\textbf{Knowledge distillation.} Knowledge distillation is a machine learning technique that seeks to compress the knowledge contained in a larger model (teacher) into a smaller one (student). It was introduced by Bucilua et al.~\cite{bucilua2006model} and later adopted to deep learning by Hinton et al.~\cite{hinton2015distilling}. Recently, Tian et al.~\cite{tian2020rethinking} show that incorporating self-distillation into FSL can boost the performance by $1-2\%$.

\textbf{Transductive inference.} Transductive inference aims to leverage the query set (which can be seen as unlabeled data during inference) in addition to the labeled support set. Transductive  approaches~\cite{liu2018learning, hou2019cross, qiao2019transductive, ziko2020laplacian, boudiaf2020transductive, dhillon2019baseline} significantly outperform their inductive counterparts, which do not exploit the query set. For example, TPN~\cite{liu2018learning} constructs a graph whose nodes are support and query images and propagates labels from the support to query images, while CAN~\cite{hou2019cross} utilizes confidently classified query images as part of the support set. Recently, TIM~\cite{boudiaf2020transductive} assumes a uniform distribution of novel classes, which is often the case for the current FSL benchmarks, and exploits that assumption for  boosting the performance. Furthermore, \cite{finn2017model, sung2018learning} can be considered as transductive methods since the information from query data is used for batch normalization.



In this paper, we improve the performance of the classifier on novel classes by introducing a novel loss function, which penalizes out-of-distribution samples. Our method is complementary to the above approaches and can be combined to establish a new state-of-the-art for FSL. As the time of camera-ready, we are aware of a concurrent work \cite{das2021importance} that also leverages the distractor samples to refine the classifier in FSL. Contrast to that work, our loss function is inherently applicable to both inductive and transductive inference. Furthermore, we also discover the effectiveness of uniform random features,  which obviates the need for accessible OOD samples for in-domain FSL.
	
\section{Preliminary}
\label{sec:preliminary}

We first define some notations used in the following sections. Let $(\mathbf{x}, y)$ consist of a sample image $\mathbf{x}$ and its corresponding class label $y$. 
Let $D_b = \{(\mathbf{x}_i, y_i) \}_{i=1}^{N_b}$ denote the labeled base samples used for feature pre-training. Next, we denote $D_s = \{(\mathbf{x}_i, y_i) \}_{i=1}^{N_s}$ and $D_q = \{(\mathbf{x}_i, y_i) \}_{i=1}^{N_q}$ as the labeled support samples and query samples respectively. Note that the labels for the query samples are only used for evaluation purposes. The support samples and query samples belong to the novel classes $C_n$, which are separated from the base classes $C_b$, \ie $C_b \cap C_n = \varnothing$.
In few-shot learning, we aim to learn a classifier that exploits support data to predict labels for query samples. We use a pre-trained feature extractor, usually kept fixed, to produce input to the classifier. 

In this work, we consider fine-tuning the classifier on the support set only (\emph{inductive learning}) and on both the support and query set (\emph{transductive learning}). 
We also consider both \emph{in-domain} and \emph{cross-domain} FSL. 
In the former, the novel classes $C_n$ and the base classes $C_b$ are from the same domain (\eg images from Image-Net), while for the latter, $C_n$ and $C_b$ are from different domains and the domains of $C_b$ and $C_n$ are referred to as the source and target domains, respectively.

\myparagraph{Pretrained feature extractor.} 
Let $f_\theta$ be the feature extractor trained on the base data with the standard cross-entropy loss, which we also refer as the \textbf{``simple baseline''}. We also seek to strengthen the baseline with orthogonal techniques such as self-supervised learning (SSL)~\cite{gidaris2019boosting, su2020does} and knowledge distillation (KD)~\cite{tian2020rethinking}. 
With SSL, we employ the \textbf{``rot baseline''}, which is trained with the standard cross-entropy loss and an auxiliary loss to predict the rotation angles of the perturbed images. We further apply the born-again strategy \cite{furlanello2018born} for the \emph{rot baseline} in two generations to construct the \textbf{``rot + KD baseline''}.
More details of our baselines can be found in Section~\ref{appendix:baseline_details}.
	
\myparagraph{Novel tasks inference.}
In the few-shot scenario, we freeze the feature extractor $f_\theta$ and train a classifier for each task. Let $\mathbf{W} = [\mathbf{w}_0, \cdots, \mathbf{w}_k]^\top \in \mathbb{R}^{K\times D}$ be the weight matrix of the classifier, where $D$ is the dimension of the encoded vector (output by the feature extractor) and $K$ is the number of classes (\ie number of ways) for each task. The predictive distribution over classes $p(k\vert \mathbf{x}_i, \mathbf{W})$ is given by:
\begin{equation}
    \label{eq:cosine_classifier}
	p(k\vert \mathbf{x}_i, \mathbf{W}) = \frac{\exp(-\gamma \cdot d(\mathbf{z}_i, \mathbf{w}_k))}{\sum_j \exp(-\gamma \cdot d(\mathbf{z}_i, \mathbf{w}_j))}, \quad \text{where} \quad \mathbf{z}_i = \frac{f_\theta(\mathbf{x}_i)}{\|f_\theta(\mathbf{x}_i)\|_2}
\end{equation}
where $\gamma$ is the learnable scaling factor~\cite{qi2018low} and $d(\cdot,\cdot)$ denotes the distance function. We use the squared Euclidean distance in our experiments unless otherwise mentioned. In our implementation, we initialize each weight vector $\mathbf{w}_k$ to the mean of the sample features in the support set $\mathbf{w}_k = \frac{1}{\vert \mathcal{S}_k\vert}\sum_{\mathbf{x} \in \mathcal{S}_k} f_\theta(\mathbf{x})$ with $\mathcal{S}_k$ being the support set of the $k^\mathrm{th}$ class similar to Prototypical Network~\cite{prototype}.

\myparagraph{Inductive bias.} Inductive learning is a process of learning a general principle by observing specific examples. 
Given limited support data, it is possible to have multiple explanations on the query data, with each corresponding to a different prediction. Inductive bias allows the learner to systematically favor one explanation over another, rather than favoring a model that overfits to the limited support data. 
Figure~\ref{fig:example} shows an ambiguous classification example that can be explained by multiple hypotheses. 
Two decision rules possibly learnt from the support data are:
1) People in suit belong to class 1; and
2) Boys belong to class 2.

Both rules are simple (satisfying Occam's razor), and can be used to classify the query sample. However, it is unclear which rule the learner would favor; it is only possible to know after training. Hence, solely learning with support data can be obscure.
One way to narrow down the hypothesis space is using counter-examples to assess the inductive validity~\cite{johnson2010mental, de2005working}.
In Figure~\ref{fig:example}, the out-of-distribution (OOD) sample hints that the suit should be considered irrelevant, and hence the rule 1 should be rejected. 


\section{Proposed approach}
We introduce our novel technique for few-shot learning namely \longalgo~(\algo).  Specifically, we attempt to regularize few-shot learning and improve the generalization of the learned prototypes by leveraging prior knowledge of in- and out-of-distribution samples. 
%
%
%
Our definition is as follows.
\textit{Positive samples} are in-distribution samples provided in the context of the current task that includes both support and query samples. 
\textit{Negative samples}, in contrast, do not belong to the context of the current task, and hence are out-of-distribution.
Negative samples can either provide additional cues that reduce ambiguity, or act as distractors to prevent the learner from overfitting. 
Note that negative samples should have the same domain as positive samples so that their cues are insightful to the learner, but positive and negative samples are not required to have the same domain as the base data. 


To effectively use positive and negative samples in few-shot learning, the following conditions must be met:
1) The regularization guided by out-of-distribution samples can be combined with traditional loss functions for classification;
2) The regularization should be applicable for both inductive and transductive inference;
and 3) 
The requirement on negative data should be minimal \ie does not need any sort of labels except for aforementioned conditions.

\subsection{Refining prototype with distractor samples}
We formulate the regularization as a new objective function for training. To capitalize negative samples to reduce the ambiguity, we propose leveraging the large-margin principle as in \emph{Large Margin Nearest Neighbors} (LMNN) \cite{weinberger2006distance, weinberger2006distance}. In these works, \etal{Weinberger} propose a loss function with two competing terms to learn a distance metric for nearest neighbor classification: a \emph{``pull''} term to penalize the large distance between the embeddings of two nearby neighbors, which likely belong to the same class, and a \emph{``push''} term to penalize the small distance between the embeddings of samples of difference classes.

In this work, we seek to learn prototypes for all categories instead of a distance metric. 
We do not have labels (\ie categories) of all samples (\eg transductive inference) but only the prior knowledge about whether a sample is \emph{in}- or \emph{out}-of-distribution. 
Thus, we adapt the above objective of margin maximization for these two groups, namely in- and out-of-distribution, with the distance function being the \textit{sum} of distances from a sample to \emph{all} prototypes. 
The goal is minimizing distances from positive samples to prototypes, while maximizing distances from negative samples to prototypes.

In our objective function, we keep the original ``pull'' term while introducing our new ``push'' term: we do not explicitly enforce large distances between positive samples and negative samples, but only attempt to maximize distances between prototypes and negative samples:
\begin{equation}
    \mathcal{L}_{naive} = \sum_{i=1}^{N_{pos}}\sum_{k=1}^K \gamma \cdot d(\mathbf{w}_k, \mathbf{x}_i) - \sum_{j=1}^{N_{neg}} \sum_{k=1}^K\gamma \cdot d(\mathbf{w}_k, \mathbf{x}_j)
\end{equation}
Note that $\gamma$ is scaling factor of distance-based classifier as in Equation \ref{eq:cosine_classifier}. However, this objective does not take into account class assignments for positive samples. To tackle this problem, we use weighted distances between prototypes and samples to simultaneously optimize both objectives:
\vspace{-5pt}
\begin{equation}
    \label{eq:loss_poodle}
    \mathcal{L}_{margin} = \underbrace{\sum_{i=1}^{N_{pos}}\sum_{k=1}^K \gamma \cdot d(\mathbf{w}_k, \mathbf{x}_i) \mathcal{S}_G[\,p(k\vert\mathbf{x}_i, \mathbf{W})]}_{\mathcal{L}_{pull}} - \underbrace{\sum_{j=1}^{N_{neg}} \sum_{k=1}^K \gamma \cdot d(\mathbf{w}_k, \mathbf{x}_j)\mathcal{S}_G[\,p(k\vert\mathbf{x}_j, \mathbf{W})]}_{\mathcal{L}_{push}}
\end{equation}
where $\mathcal{S}_G[\cdot]$ is the stop-gradient operator. Intuitively, the positive sample will \emph{``pull''} the prototypes to its location proportional to its distance to the prototypes. Subsequently, the prototype of each class will move closer to the positive samples of that class. At the same time, the prototype is enforced to move away from the negative samples, 
thus discarding features that might lead to high similarity to out-of-distribution data. 

We note that 
removing $\mathcal{S}_G[\cdot]$ in Equation~\ref{eq:loss_poodle} would result in a different underlying objective, which empirically leads to a decrease in performance. Particularly, the objective without stop-gradient will also be compounded of an auxiliary term for entropy maximization (see Section \ref{sec:on_stop_gradient_operator}). Thus, devoid of meticulous regularization would deteriorate the performance. The above observation is in line with \etal{Boudiaf}~\cite{boudiaf2020transductive}.
In summary, \algo optimizes the classifier on novel tasks with the following objectives:
\begin{equation}
    \boxed{
        \mathcal{L}_{\algo} = \mathcal{L}_{ce} + \alpha\cdot\mathcal{L}_{pull} - \beta\cdot\mathcal{L}_{push}
    }
\end{equation}




where $\alpha$ and $\beta$ control the ``push'' and ``pull'' coefficients respectively and $\mathcal{L}_{ce}$ denotes the standard cross-entropy loss. To the best of our knowledge, the proposed objective is the first loss function (in test phase) that can be effective for both inductive and transductive inference. Please see the supplemental document for the pseudo-code (Section \ref{appendix:pseudo_code}).

\subsection{On negative samples}
\label{sec:discussion}
\myparagraph{Choice of negative samples.} 
For \textit{in-domain} FSL, we simply leverage the base data as negative samples.
For \textit{cross-domain} FSL, one approach would be using a set of samples drawn from classes that are disjoint to novel classes $C_n$ as negative examples. However, in some cases such negative examples might not be available. Thus, we consider another approach where we have a set of unlabeled data of a set of classes $C_u$ from the target domain ($C_u$ might overlap with $C_n$) as negative examples, similar to~\cite{phoo2021selftraining}. We name the above two approaches as \emph{disjoint} and \emph{noisy} negative sampling, respectively in the context of cross-domain FSL.
Intuitively, when the number of categories of unlabeled data (\ie $\vert C_u \vert$) grow larger, the noise of mixing positive and negative samples in \emph{noisy} negative samples pool will be reduced.

\myparagraph{The burden of additional data.} It is worth pointing out that \algo does not break the setup of FSL as it does not require any additional data of novel classes, thus, still has a few samples from novel tasks. As \algo obligates additional data in the form of OOD samples, one might ask how practical the algorithm is. For in-domain FSL, the negative samples can be easily obtained by adopting the training data as we already know $C_n \cap C_b = \varnothing$. Even in extreme case where we only have access to the pretrained models, we empirically find that ($\ell_2$-normalized) uniformly random noise can work surprisingly well (Section \ref{subsec:random_features}). For cross-domain FSL, even though \algo requires additional OOD data, we find that these samples can be: 1) noisy with both positive and negative samples; and 2) fairly efficient - in our experiments, we can ``reuse" $400$ OOD samples for all tasks, which is very efficient compared to other methods that might use up to $20\%$ \emph{unlabeled} data of training set \cite{phoo2021selftraining}.

\subsection{Intriguing effectiveness of uniform random features}
\label{subsec:random_features}
One caveat of \algo is the obligation of accessing to OOD samples. Fortunately, we find that \algo can use the uniform random features on the hypersphere as negative samples \textbf{for in-domain FSL}. Particularly, we can uniformly sample latent vectors from $\mathbb{S}^{D-1}$ as an alternative to the distractor samples from base training data, thus, eliminating the need for accessing to training data.

Our use of uniform distribution as negative examples is inspired by that feature uniformity is a desirable property for contrastive loss~\cite{wang2020understanding, chen2020intriguing}, and so a good representation prefers such uniformity. The problem of uniformly distributing points on the unit hypersphere is related to minimizing pairwise loss~\cite{kallenberg1997foundations, borodachov2019discrete}, which links well to the theory of supervised classification with softmax and cross-entropy (equivalent to minimizing pairwise loss or maximizing mutual information)~\cite{boudiaf2020unifying, qin2019rethinking}.

Therefore, using uniform distribution as negative samples works for in-domain FSL because they will approximate the features of samples from base training data. For cross-domain FSL, this approach will fail because random features (which are similar to samples from the training domain) have a large discrepancy with the target domain, thus, not inducing meaningful cues. Intuitively, the more similar between domains of OOD and test samples, the higher performance gain \algo can achieve. Our experiments in Section \ref{subsec:ablation} empirically prove aforementioned postulation.

\section{Experiments}
\label{sec:experiments}
In this section, we conduct extensive experiments to demonstrate the performance gain of our method on standard inductive, transductive, cross-domain FSL. To demonstrate the robustness of our method across datasets/network architectures, \textbf{we keep the hyperparameters fixed for all experiments.}

\subsection{Experimental setup}
\myparagraph{Datasets.} We evaluate our approach on three common FSL datasets. The \textit{mini}-Imagenet dataset \cite{matching_net} consists of 100 classes chosen from the ImageNet dataset \cite{imagenet} including 64 training, 16 validation, and 20 test classes with 60,000 images of size 84 $\times$ 84. 
The \textit{tiered}-Imagenet \cite{tiered_imagenet} is another FSL dataset which is also derived from the ImageNet dataset with 351 base, 97 validation, and 160 test classes with 779,165 images of size 84 $\times$ 84. 
Caltech-UCSD Birds (CUB) has 200 classes split into 100, 50, 50 classes for train, validation and test following \cite{chen2019closer}.
Furthermore, we also carry out experiments on iNaturalist 2017 (iNat) \cite{van2018inaturalist}, EuroSAT \cite{helber2019eurosat}, and ISIC-2018 (ISIC) \cite{codella2019skin} for the (extreme) cross-domain FSL. The description for these datasets can be found in the supplemental document (Section \ref{appendix:dataset}).

\myparagraph{Implementation details.}
We use ResNet12 as our feature extractor. It is a residual network \cite{ye2020few} with 12 layers split into 4 residual blocks.
For pre-training on the base classes, we train our backbones with the standard cross-entropy loss for 100 epochs. The optimizer has a weight decay of $5e^{-4}$, and the initial learning rate of 0.05 is decreased by a factor of 10 after 60, 80 epochs in mini-ImageNet and 60, 80, 90 epochs in tiered-ImageNet. We use the batch size of 64 for all the networks.
For fine-tuning on the novel classes, we utilize Adam optimizer \cite{kingma2014adam} with fixed learning rate of $0.001$, $\beta_1 = 0.9$, $\beta_2 = 0.999$, and do not use weight decay. The classifier is trained with 250 iterations. The coefficients of ``push/pull'' loss are $\alpha=1$ and $\beta=0.5$ respectively. For negative samples, we randomly select $K=400$ samples from the out-of-distribution samples pool for each novel task. We evaluate the performance of \algo in 5-way-1-shot and 5-way-5-shot settings on 2000 random tasks with 15 queries each.

\begin{table}
	\centering
    	\caption{Comparison to different baselines for the standard FSL (a.k.a in-domain FSL) on \textit{mini}-ImageNet, \textit{tiered}-Imagenet and CUB in the \textbf{inductive} setting with \emph{Resnet-12} as backbone. Here, \algo-B and \algo-R indicate that the negative samples are sampled from the base classes and the random uniform distribution respectively.}
	    \begin{tabular}{p{1.9cm}p{1.9cm}p{1.1cm}p{1.1cm}p{1.1cm}p{1.1cm}p{1.1cm}p{1.1cm}}
		\toprule
		& &\multicolumn{2}{c}{\textbf{\textit{mini}-ImageNet}} & \multicolumn{2}{c}{\textbf{\textit{tiered}-ImageNet}} & \multicolumn{2}{c}{\textbf{CUB}} \\
		Baseline & Variant & 1-shot & 5-shot & 1-shot & 5-shot & 1-shot & 5-shot\\
		\toprule
		\multirow{3}{*}{Simple} & CE Loss & 61.33 & 80.76 &67.42 & 84.44 & 75.60 & 89.88 \\
		& \cellcolor{pearDark!20}\algo-B & \cellcolor{pearDark!20}63.79 & \cellcolor{pearDark!20}\textbf{81.41} & \cellcolor{pearDark!20}69.26 & \cellcolor{pearDark!20}84.97 & \cellcolor{pearDark!20}75.96 & \cellcolor{pearDark!20}\textbf{90.06} \\
		& \cellcolor{pearDark!20}\algo-R & \cellcolor{pearDark!20}\textbf{64.38} &  \cellcolor{pearDark!20}81.35 & \cellcolor{pearDark!20}\textbf{69.34} & \cellcolor{pearDark!20}\textbf{85.03} & \cellcolor{pearDark!20}\textbf{76.22} & \cellcolor{pearDark!20}89.99 \\
		
		\midrule
		\multirow{3}{*}{Rot}& CE Loss & 64.57 & 82.89 & 68.26 & 85.09 & 76.39 & 91.10 \\
		& \cellcolor{pearDark!20}\algo-B & \cellcolor{pearDark!20}66.78 & \cellcolor{pearDark!20}\textbf{83.65} & \cellcolor{pearDark!20}69.74 & \cellcolor{pearDark!20}85.45 & \cellcolor{pearDark!20}77.26 & \cellcolor{pearDark!20}91.30 \\
		& \cellcolor{pearDark!20}\algo-R & \cellcolor{pearDark!20}\textbf{67.20} & \cellcolor{pearDark!20}83.54 & \cellcolor{pearDark!20}\textbf{69.86} & \cellcolor{pearDark!20}\textbf{85.56} & \cellcolor{pearDark!20}\textbf{77.35} & \cellcolor{pearDark!20}\textbf{91.43} \\

		\midrule
		\multirow{3}{*}{Rot + KD} & CE Loss & 65.91 & 82.95 & 69.43 & 84.93 & 79.70 & 92.28 \\
		& \cellcolor{pearDark!20}\algo-B & \cellcolor{pearDark!20}67.50 & \cellcolor{pearDark!20}\textbf{83.71} & \cellcolor{pearDark!20}70.42 & \cellcolor{pearDark!20}\textbf{85.26} & \cellcolor{pearDark!20}\textbf{80.23} & \cellcolor{pearDark!20}92.36 \\
		& \cellcolor{pearDark!20}\algo-R & \cellcolor{pearDark!20}\textbf{67.80} & \cellcolor{pearDark!20}83.50 & \cellcolor{pearDark!20}\textbf{70.47} & \cellcolor{pearDark!20}85.24 & \cellcolor{pearDark!20}80.05 & \cellcolor{pearDark!20}\textbf{92.37} \\
		\bottomrule
    	\end{tabular}
	\label{tab:compared_baseline_inductive}
\end{table}

\begin{table}[t]
	\small
	\centering
	\caption{Comparison to different baselines on \textit{mini}-ImageNet, \textit{tiered}-Imagenet and CUB with and without the proposed loss in the \textbf{transductive} settings. Here, $\mathcal{L}_{pull}^b$ and $\mathcal{L}_{pull}^u$ indicate that the negative samples are sampled from the base classes and the random uniform distribution respectively.}
	\resizebox{\textwidth}{!}{
	    \begin{tabular}{ccccccccc}
	        \toprule
	    	& & \multicolumn{2}{c}{\textbf{\textit{mini}-ImageNet}} &     \multicolumn{2}{c}{\textbf{\textit{tiered}-ImageNet}} & 						    \multicolumn{2}{c}{\textbf{CUB}}\\
	    	Baseline & Loss & 1-shot & 5-shot & 1-shot & 5-shot & 1-shot & 5-shot \\
	    	\toprule
	    	\multirow{4}{*}{Simple} & $\mathrm{CE}$ & 61.33 & 80.76 & 67.42 & 84.44 & 75.60 & 89.88\\
	    	& $\mathrm{CE} + \mathcal{L}_{pull}$ & 70.91 & 83.04 & 75.58 & 86.10 & 84.91 & 91.60 \\
	    	& $\mathrm{CE} + \mathcal{L}_{pull} - \mathcal{L}_{push}^{b}$ & 74.21 & \textbf{83.71} & 78.72 & 86.57 & 87.04 & \textbf{91.84} \\
	    	& $\mathrm{CE} + \mathcal{L}_{pull} - \mathcal{L}_{push}^{u}$ & \textbf{74.82} & 83.67 & \textbf{79.24} & \textbf{86.64} & \textbf{87.13} & 91.60 \\
	    	
	    	\midrule
	    	\multirow{4}{*}{Rot} & $\mathrm{CE}$ & 64.57 & 82.89 & 68.26 & 85.09 & 76.39 & 91.10\\
	    	& $\mathrm{CE} + \mathcal{L}_{pull}$ & 74.42 & 85.20 & 76.59 & 86.75 & 85.93 & 93.02 \\
	    	& $\mathrm{CE} + \mathcal{L}_{pull} - \mathcal{L}_{push}^{b}$ & 77.30 & \textbf{85.91} & 79.74 & 87.25 & 88.68 & 93.23 \\
	    	& $\mathrm{CE} + \mathcal{L}_{pull} - \mathcal{L}_{push}^{u}$ & \textbf{77.69} & 85.86 & \textbf{80.31} & \textbf{87.33} & \textbf{89.07} & \textbf{93.35} \\
	    	
	    	\midrule
	    	\multirow{4}{*}{Rot + KD} & $\mathrm{CE}$ & 65.91 & 82.95 & 69.43 & 84.93 & 79.70 & 92.28\\
	    	& $\mathrm{CE} + \mathcal{L}_{pull}$ & 74.90 & 85.09 & 77.07 & 86.50 & 88.22 & 93.68 \\
	    	& $\mathrm{CE} + \mathcal{L}_{pull} - \mathcal{L}_{push}^{b}$ & \textbf{77.56} & \textbf{85.81} & 79.67 & \textbf{86.96} & 89.88 & \textbf{93.80} \\
	    	& $\mathrm{CE} + \mathcal{L}_{pull} - \mathcal{L}_{push}^{u}$ & 77.24 & 85.53 & \textbf{79.97} & 86.91 & \textbf{90.02} & 93.68 \\
	    	\bottomrule
	    \end{tabular}
	}
	\label{tab:compare_baseline_transductive}
\end{table}

\subsection{Standard FSL}
We first experiment with the standard FSL setup (also known as in-domain FSL) in which the number of query samples is uniformly distributed among classes.

\textbf{Evaluation with various baselines.} Table~\ref{tab:compared_baseline_inductive} shows the results of our approach with various baselines, which described in Section \ref{sec:preliminary}, with the inductive inference (positive samples are support images). As can be seen, our approach consistently boosts the performance of all baselines by a large margin (1-3$\%$). Interestingly, the \algo-R variant outperforms the CE loss variant significantly and performs comparatively with the \algo-B variant. We hypothesize that the normalized representations of the samples extracted from the base classes are uniformly distributed.

Table \ref{tab:compare_baseline_transductive} demonstrates the efficacy of each loss term of \algo in transductive inference (positive samples are support and query images). We can see that using the ``pull'' loss with query samples improve the inductive baseline significantly, being as effective as other transductive algorithms. Combining with ``push'' term, the classifier is further enhanced.

\begin{figure}[t]
	\centering
	\begin{subfigure}[b]{0.5\textwidth}
		\centering
		\includegraphics[width=\linewidth]{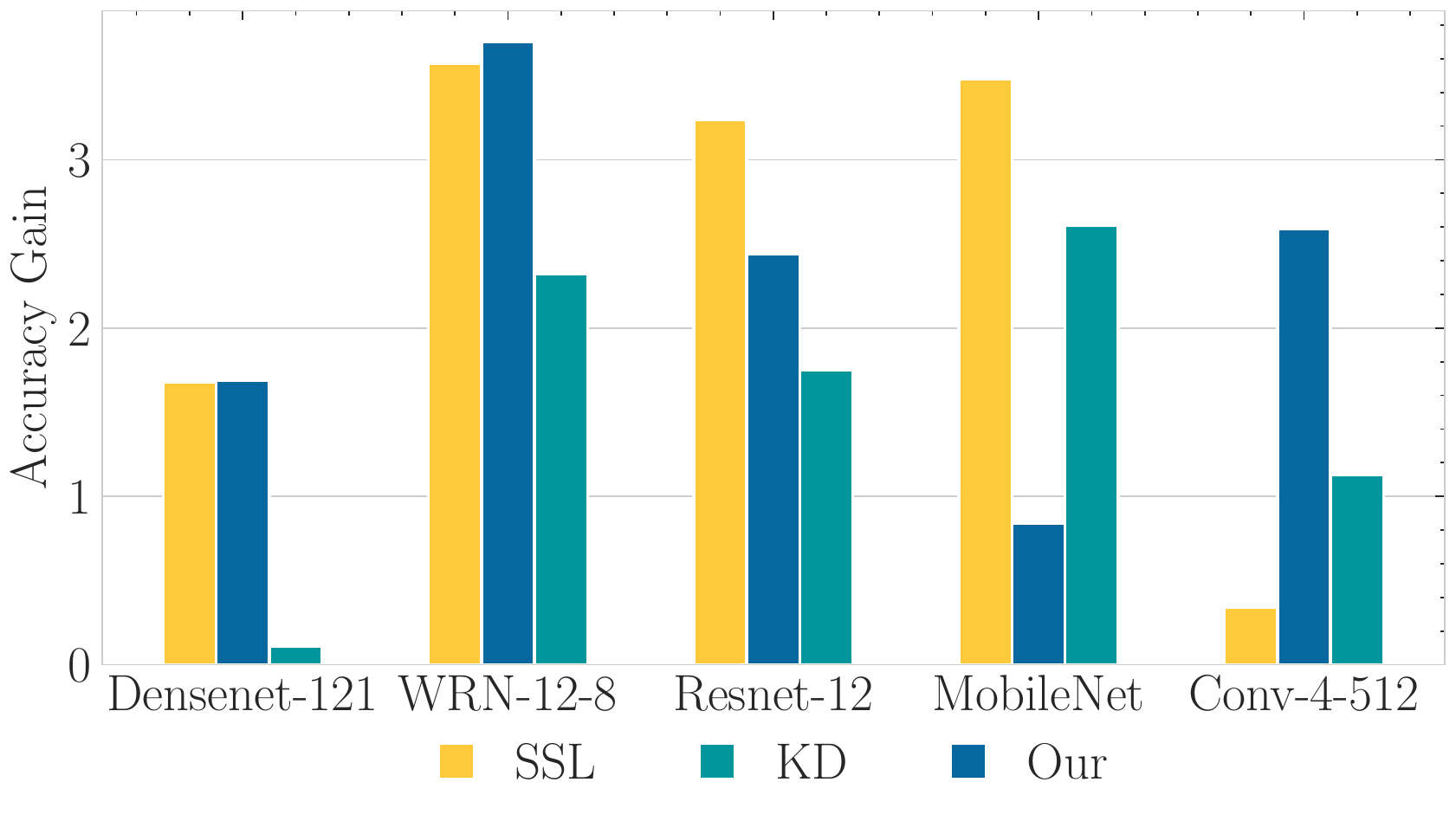}
		\caption{}
		\label{fig:compare_ssl_kd_a}
	\end{subfigure}
	\hfill
	\begin{subfigure}[b]{0.45\textwidth}
		\centering
		\includegraphics[width=\linewidth]{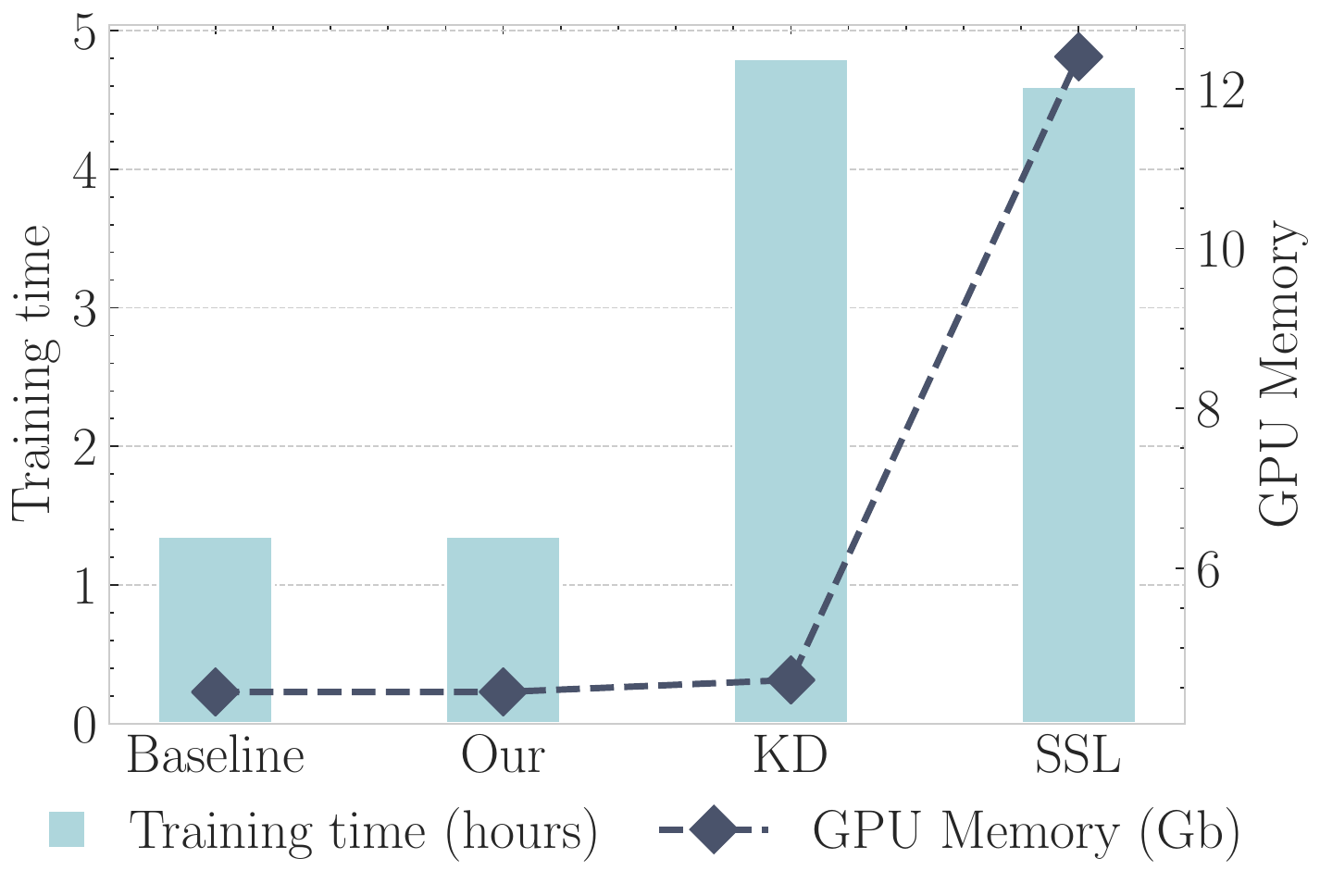}
		\caption{}
		\label{fig:compare_ssl_kd_b}
	\end{subfigure}
	\caption{(a) Effectiveness of our approach applied to standard FSL. (a) Our approach consistently yields accuracy gains on different backbones, in comparison with those obtained by self-supervised loss (SSL) and knowledge distillation (KD) in 5-way-1-shot protocol on \textit{mini}-Imagenet with \textbf{inductive} setting. (b) Total time and memory cost (in training stage) when adopting SSL, KD, and our method. Note that we apply KD for $T=2$ generation as in \cite{tian2020rethinking}. No large overhead is incurred in our method.}
	\label{fig:compare_ssl_kd}
	\vspace{-5pt}
\end{figure}

We also conduct experiments with additional loss functions including self-supervised loss (SSL) and knowledge distillation (KD) in Figure \ref{fig:compare_ssl_kd_a}. Our approach consistently improves the generalization of all baselines without additional computation cost (in the training phase) as justified in Figure \ref{fig:compare_ssl_kd_b}. The improvement of \algo on other backbones and datasets is reported in the supplemental document (Section \ref{appendix:other_backbones}).

\myparagraph{Comparison to the state-of-the-art approaches.}\label{sec:benchmark}
We report the performance of our network in comparison with state-of-the-art methods in both transductive and inductive settings (with and without information from the query images) in Table \ref{tab:benchmark_results}. We can see that our approach remarkably improves the performance of the baseline and achieves a comparable performance with the state-of-the-art approaches in the tiered-ImageNet. In mini-ImageNet and CUB we significantly outperform the prior work in both inductive and transductive settings. Experimental results on other backbones are reported in the supplementary document (Section \ref{appendix:other_backbones}).

\begin{table}
	\centering
	\caption{Comparison to the state-of-the-art methods on \textit{mini}-ImageNet, \textit{tiered}-Imagenet and CUB using \textbf{inductive} and \textbf{transductive} settings. The results obtained by our models (\textcolor{pearDark}{blue pearl}-shaded) are averaged over 2,000 episodes.}
	\resizebox{\textwidth}{!}{\begin{tabular}{lcccccccc}
		\toprule
		& & &\multicolumn{2}{c}{\textbf{\textit{mini}-ImageNet}} & \multicolumn{2}{c}{\textbf{\textit{tiered}-ImageNet}} & \multicolumn{2}{c}{\textbf{CUB}} \\
		Method & Transd. & Backbone & 1-shot & 5-shot & 1-shot & 5-shot & 1-shot & 5-shot\\
		\toprule
		MAML \cite{finn2017model}& \multirow{9}{*}{\xmark} & ResNet-18 & 49.6 & 65.7 & - & - & 68.4 & 83.5 \\
         RelatNet \cite{hou2019cross}& & ResNet-18 & 52.5 & 69.8 &  - & - & 68.6 & 84.0 \\
         MatchNet \cite{matching_net}& & ResNet-18 & 52.9 & 68.9 &  - & - & 73.5 & 84.5 \\
         ProtoNet \cite{prototypical_nets}& & ResNet-18 & 54.2 & 73.4 &  - & - & 73.0 & 86.6 \\
         Neg-cosine \cite{liu2020negative}& & ResNet-18 & 62.3 & 80.9 & - & - & 72.7 & 89.4 \\
         MetaOpt \cite{lee2019meta} & & ResNet-12 & 62.6 & 78.6 & 66.0 & 81.6 & - & -\\
         SimpleShot \cite{simpleshot} & & ResNet-18 & 62.9 & 80.0 & 68.9 & 84.6 & 68.9 & 84.0 \\
         Distill \cite{tian2020rethinking} & & ResNet-12 & 64.8 & 82.1 & \textbf{71.5} & \textbf{86.0} & - & -\\
		\rowcolor{pearDark!20}Rot + KD + \algo & & ResNet-12 & \textbf{67.80} & \textbf{83.72} & 70.42 & 85.26 & \textbf{80.23} & \textbf{92.36} \\

        \midrule
		RelatNet + T \cite{hou2019cross} & \multirow{8}{*}{\cmark} & ResNet-12 & 52.4 & 65.4 & - & - & - & -\\
		TPN \cite{liu2018learning} & & ResNet-12 & 59.5 & 75.7 & - & - & - & -\\
		TEAM \cite{qiao2019transductive}& &  ResNet-18 & 60.1 & 75.9 & - & - & - & -\\
		Ent-min \cite{dhillon2019baseline}& & ResNet-12 & 62.4 & 74.5 & 68.4 & 83.4 & - & -\\
		CAN+T \cite{hou2019cross}& &  ResNet-12 & 67.2 & 80.6 & 73.2 & 84.9 & - & -\\
		LaplacianShot \cite{ziko2020laplacian} & & ResNet-18 & 72.1 & 82.3 & 79.0 & 86.4 & 81.0 & 88.7 \\
		TIM-GD \cite{boudiaf2020transductive} & & ResNet-18 & 73.9 & 85.0 & \textbf{79.9} & \textbf{88.5} & 82.2 & 90.8 \\
		\rowcolor{pearDark!20} Simple + \algo & & ResNet-12 & 74.21 & 83.71 & 78.72 & 86.57 & 87.04 & 91.84 \\
		\rowcolor{pearDark!20}Rot + KD + \algo & & ResNet-12 & \textbf{77.56} & \textbf{85.81} & 79.67 & 86.96 & \textbf{89.93} & \textbf{93.78} \\
    	\bottomrule
	\end{tabular}}
	\label{tab:benchmark_results}
\end{table}

\subsection{Cross-domain FSL}   
In this section, we conduct experiments to demonstrate the efficacy of our approach even with a challenging task: \textbf{extreme} cross-domain FSL as first introduced in \cite{guo2020broader, phoo2021selftraining}. Many FSL algorithms are known to fail in such a challenging scenario~\cite{chen2019closer}.

Recall that we have two setups for cross-domain FSL (Section \ref{sec:discussion}), in which cases, the 
negative samples are drawn from \emph{disjoint} and \emph{noisy} negative samples pool.
Here, we provide the detailed setting of each scheme when transferring a network trained on \emph{mini}-Imagenet to target domains. It is worth mentioning that we only consider \emph{inductive} inference for comparison with prior work.

\textbf{Disjoint negative samples.} As mentioned before, we evaluate the results of cross-domain FSL on test split of these datasets while employing the train split to draw negative samples. Since the number of categories in EuroSAT and ISIC is relatively small (10 and 7 respectively) compare to the number of ``way'' in novel task (5), we only concern with iNat and CUB. Precisely, we follow \cite{chen2019closer} and \cite{simpleshot} to split CUB and iNat respectively. 

\textbf{Noisy negative samples.} We assume that we have $20\%$ unlabeled data from the target domain and the rest $80\%$ of data is used for testing. Although a large number of classes are beneficial to \algo as discussed above, we also carry out experiments with ISIC and EuroSAT to evaluate the performance of \textbf{extreme} cross-domain FSL with our approach.

From the Table \ref{tab:domin_shift_results} we can see that the improvement of \algo with the \textbf{disjoint} negative samples (the bottom rows) and the \textbf{noisy} negative samples (the middle rows) when transferring knowledge from mini-ImageNet to iNat, CUB, ISIC, and EuroSAT.
We can observe that despite the extreme gap between target/source domains and a very noisy mix of in- and out-of-distribution samples in negative samples pool (in case of EuroSAT and ISIC), \algo can successfully boost the performance of baselines in all experiments. We also report the performance of other approaches, which have same setup as \textbf{noisy} negative samples, in Table \ref{tab:domin_shift_results}.

\begin{table}
	\newlength\wexp
	\settowidth{\wexp}{\textbf{\textit{mini}-ImageNet} $\rightarrow$ \textbf{CUB}}
	\centering
	\small
	\caption{The results of the domain-shift setting from \textit{mini}-Imagenet to CUB, iNat, ISIC, and EuroSAT with \emph{Rot + KD baseline}. The results obtained by our models (\textcolor{pearDark}{blue pearl}-shaded) are averaged over 2,000 episodes. The baselines with $\star$ notation show the results using the setup of \textbf{disjoint} negative samples, and the baselines without the $\star$ show the results using the setup of \textbf{noisy} negative samples.}
	\resizebox{\textwidth}{!}{\begin{tabular}{lcccccccc}
	\toprule
	& \multicolumn{2}{c}{\textbf{CUB}} & \multicolumn{2}{c}{\textbf{iNat}} &\multicolumn{2}{c}{\textbf{ISIC}} & \multicolumn{2}{c}{\textbf{EuroSAT}} \\
	Baseline & 1-shot & 5-shot & 1-shot & 5-shot & 1-shot & 5-shot & 1-shot & 5-shot\\
	\toprule
	Transfer \cite{phoo2021selftraining} & - & - & - & - & 30.71 & 43.08 & 60.73 & 80.30 \\
	SimCLR \cite{phoo2021selftraining} & - & - & - & - & 26.25 & 36.09 & 43.52 & 59.05 \\
	Transfer + SimCLR \cite{phoo2021selftraining} & - & - & - & - & 32.63 & 45.96 & 57.18 & 77.61 \\
	STARTUP (no SS) \cite{phoo2021selftraining} & - & - & - & - & 32.24 & 46.48 & 62.90 & 81.81 \\
	STARTUP \cite{phoo2021selftraining} & - & - & - & - & 32.66 & \textbf{47.22} & 63.88 & \textbf{82.29} \\
	Rot + KD + CE & 49.98 & 69.44 & 48.90 & 66.58 & 32.35 & 43.81 & 65.00 & 80.01 \\
	\rowcolor{pearDark!20}Rot + KD + \algo-R & 50.04 & 69.78 & 48.94 & 66.87 & 32.21 & 43.87 & 63.97 & 79.85 \\
	\rowcolor{pearDark!20}Rot + KD + \algo-B & \textbf{52.61} & \textbf{70.78} & \textbf{50.62} & \textbf{67.31} &\textbf{33.56} & 44.17 & \textbf{66.21} & 80.51 \\
	\midrule
	Rot + KD + CE $^{\star}$ & 50.28 & 69.64 & 48.80 & 66.72 & - & - & - & - \\
	\rowcolor{pearDark!20}Rot + KD + \algo-R $^{\star}$ & 50.25 & 69.97 & 48.84 & 67.03 & - & - & - & - \\
	\rowcolor{pearDark!20}Rot + KD + \algo-B $^{\star}$ & \textbf{53.11} & \textbf{70.96} & \textbf{50.46} & \textbf{67.45} & - & - & - & - \\
	\bottomrule
	\end{tabular}}
	\label{tab:domin_shift_results}
\end{table}

\subsection{Ablation study}
\label{subsec:ablation}
In this section, we present results of some presentative ablation study to get more insight of \algo behaviors. For more in-depth experiments, we refer reader to supplementary document.

\subsubsection{The effect of domain discrepancy between positive-negative samples}
Intuitively, the more similar between domains of OOD and test samples, the higher performance gain POODLE can achieve. 
To understand how our method works when OOD data is from various domains, we train our classifier and compare the performance when using random uniform distribution and other datasets as negative examples. The result of this experiment is reported in Table \ref{tab:test_ood_domains_ablation}. We can observer that the more similar of OOD domain to test domain, the higher the performance gain we can achieve. Thus, we should always aim to leverage the OOD samples of test domain.

\begin{table}
    \caption{Evaluating (simple) Resnet-12 trained on mini-Imagenet on different test/OOD domains in the 1-shot inductive protocol (10,000 episodes). \textit{random, mini, tiered, CUB, EuroSAT} denote \textit{mini}-Imagenet, \textit{tiered}-Imagenet, CUB, EuroSAT dataset respectively. The OOD/test samples are drawn from the standard train/test split of each dataset, respectively. The 95\% confidence interval is roughly 0.20 for all results.}
    \begin{tabular}{llrrrrrr}
        \toprule
        &                 & \multicolumn{6}{c}{\textbf{OOD domain}} \\ 
        &                 & \textit{w/o OOD} & \textit{random} & \textit{mini} & \textit{tiered} & \textit{CUB}   & \textit{EuroSAT} \\ 
        
        \midrule
        \multirow{4}{*}{\textbf{Test domain}} & \textit{mini}   & $61.63$          & $\mathbf{64.30}$          & $64.08$         & $63.72$           & $62.71$ & $61.76$   \\
        & \textit{tiered} & $63.04$             & $64.28$          & $64.01$         & $\mathbf{64.50}$           & $63.85$ & $62.78$   \\
        & \textit{CUB}             & $48.55$             & $48.88$          & $49.01$         & $49.10$           & $\mathbf{51.40}$ & $49.18$   \\
        & \textit{EuroSAT}         & $65.18$             & $63.85$          & $64.58$         & $64.25$           & $64.70$ & $\mathbf{66.04}$   \\ \bottomrule
    \end{tabular}
    \label{tab:test_ood_domains_ablation}
\end{table}
\paragraph{On the performance of random features:} For in-domain FSL \ie test domains are mini-Imagenet and tiered-Imagenet, using uniform examples has the best and second-best accuracy when tested on mini-ImageNet and tiered-ImageNet, respectively. As aforesaid, the uniform random features will reflect the distribution of mini-Imagenet samples (which the network is pre-trained on). Thus, random features are effective for in-domain FSL.

For cross-domain FSL (\ie test domains are CUB and EuroSAT), using random uniform features - which is approximately equal to using samples from mini-Imagenetet - does not work because the discrepancy between the source and target domain is extremely large, \eg animal (mini-Imagenet) vs satellite images (EuroSAT). 

\subsubsection{Comparison against simple approaches for enhance pretrained models}
\begin{table}
	\centering
	\small
	\caption{The results of the various approaches to learn a better classifier with pretrained \emph{Rot + KD baseline} on \textit{mini}-Imagenet. The results are obtained by averaging over 10,000 episodes.}
	\begin{tabular}{lcc}
	\toprule
	Method & 1-shot & 5-shot \\
	\toprule
	baseline & $66.32\pm0.20$ & $82.99\pm0.13$\\
	1) w/ learning cosine classifier & $66.66\pm0.20$ & $82.84\pm0.13$\\
	2) w/ naive OOD & $66.36\pm0.20$ & $83.08\pm0.13$ \\
	3) w/ large-margin w/o negative samples & $66.32\pm0.20$ & $83.01\pm0.13$ \\
	4) w/ label smoothing & $66.13\pm0.20$ & $80.81\pm0.13$ \\
	\rowcolor{pearDark!20}\algo-B & $\mathbf{67.84\pm0.20}$ & $\mathbf{83.72\pm0.13}$ \\
	\rowcolor{pearDark!20}\algo-R & $\mathbf{68.20\pm0.20}$ & $\mathbf{83.60\pm0.13}$ \\
	\bottomrule
	\end{tabular}
	\label{tab:compare_simple_approaches}
\end{table}

To better understand the effectiveness of our method, we also compare to four methods that learn better classifiers on novel classes with pre-trained features. The results are reported in Table \ref{tab:compare_simple_approaches}.
\begin{enumerate}[leftmargin=*]
    \item \emph{Learning cosine classifier}: The classifier in inference phase is fine-tuned using CE loss with the support samples of each task. In our experiments, this approach does not bring any meaningful improvement similar to \cite{simpleshot}.
    \item \emph{Naive OOD}: we fine-tune a $(k+1)$-way classifier with 1 additional class for OOD samples using CE loss. Naive OOD performs worse because the OOD samples are drawn from several classes of the training set, which are well-clustered and separated, we cannot find a "prototype" with a linear classifier to match all of them.
    \item \emph{Large-margin w/o negative samples}: only use push term in \algo. This approach is not better than cosine-distance because the initialized prototype (mean of all samples from specific class) is already well-clustered and optimal \ie close to the ground-truth class prototype and far away from others for the observed samples.
    \item \emph{Label smoothing:} we use label smoothing for CE loss (ground-truth class has the probability of 0.9 and uniformly distribute 0.1 to the rest. This method does not work well because it only makes learned prototypes not be far away from samples from other classes.
\end{enumerate}
\section{Conclusions and future work}
\label{sec:conclusion}
In this work, we have proposed the concept of leveraging out-of-distribution samples set to improve the generalization of few-shot learners and realize it by a simple yet effective objective function. Our approach consistently boosts the performance of FSL across different backbone networks, inference types (inductive/transductive), and the challenging cross-domain FSL.

Future work might seek to exploit different sampling strategies (\ie how to select negative samples) to further boost the performance and reduce time/memory complexity; another interesting direction is enhancing the robustness of the classifier when we have both positive and negative samples in the same sampling pool; leveraging domain adaptation to reduce the need of in-domain negative samples is also a promising research direction.


\bibliographystyle{plain}
\bibliography{main}

\begin{thebibliography}{10}

\bibitem{borodachov2019discrete}
Sergiy~V Borodachov, Douglas~P Hardin, and Edward~B Saff.
\newblock {\em Discrete energy on rectifiable sets}.
\newblock Springer, 2019.

\bibitem{boudiaf2020transductive}
Malik Boudiaf, Ziko~Imtiaz Masud, J{\'e}r{\^o}me Rony, Jos{\'e} Dolz, Pablo
  Piantanida, and Ismail~Ben Ayed.
\newblock Transductive information maximization for few-shot learning.
\newblock {\em arXiv preprint arXiv:2008.11297}, 2020.

\bibitem{boudiaf2020unifying}
Malik Boudiaf, J{\'e}r{\^o}me Rony, Imtiaz~Masud Ziko, Eric Granger, Marco
  Pedersoli, Pablo Piantanida, and Ismail~Ben Ayed.
\newblock A unifying mutual information view of metric learning: cross-entropy
  vs. pairwise losses.
\newblock In {\em European Conference on Computer Vision}, pages 548--564.
  Springer, 2020.

\bibitem{bucilua2006model}
Cristian Buciluǎ, Rich Caruana, and Alexandru Niculescu-Mizil.
\newblock Model compression.
\newblock In {\em Proceedings of ACM SIGKDD International Conference on
  Knowledge Discovery and Data Mining}, pages 535--541, 2006.

\bibitem{chen2020intriguing}
Ting Chen and Lala Li.
\newblock Intriguing properties of contrastive losses.
\newblock {\em arXiv preprint arXiv:2011.02803}, 2020.

\bibitem{chen2019closer}
Wei-Yu Chen, Yen-Cheng Liu, Zsolt Kira, Yu-Chiang~Frank Wang, and Jia-Bin
  Huang.
\newblock A closer look at few-shot classification.
\newblock {\em arXiv preprint arXiv:1904.04232}, 2019.

\bibitem{codella2019skin}
Noel Codella, Veronica Rotemberg, Philipp Tschandl, M~Emre Celebi, Stephen
  Dusza, David Gutman, Brian Helba, Aadi Kalloo, Konstantinos Liopyris, Michael
  Marchetti, et~al.
\newblock Skin lesion analysis toward melanoma detection 2018: A challenge
  hosted by the international skin imaging collaboration (isic).
\newblock {\em arXiv preprint arXiv:1902.03368}, 2019.

\bibitem{das2021importance}
Rajshekhar Das, Yu-Xiong Wang, and Jose~MF Moura.
\newblock On the importance of distractors for few-shot classification.
\newblock In {\em Proceedings of the IEEE/CVF International Conference on
  Computer Vision}, pages 9030--9040, 2021.

\bibitem{de2005working}
Wim De~Neys, Walter Schaeken, and G{\'e}ry d'Ydewalle.
\newblock Working memory and everyday conditional reasoning: Retrieval and
  inhibition of stored counterexamples.
\newblock {\em Thinking \& Reasoning}, 11(4):349--381, 2005.

\bibitem{dhillon2019baseline}
Guneet~S Dhillon, Pratik Chaudhari, Avinash Ravichandran, and Stefano Soatto.
\newblock A baseline for few-shot image classification.
\newblock {\em arXiv preprint arXiv:1909.02729}, 2019.

\bibitem{doersch2020crosstransformers}
Carl Doersch, Ankush Gupta, and Andrew Zisserman.
\newblock Crosstransformers: spatially-aware few-shot transfer.
\newblock {\em arXiv preprint arXiv:2007.11498}, 2020.

\bibitem{doersch2017multi}
Carl Doersch and Andrew Zisserman.
\newblock Multi-task self-supervised visual learning.
\newblock In {\em Proceedings of the IEEE International Conference on Computer
  Vision}, pages 2051--2060, 2017.

\bibitem{edgington1995conditionals}
Dorothy Edgington.
\newblock On conditionals.
\newblock {\em Mind}, 104(414):235--329, 1995.

\bibitem{feinman2018learning}
Reuben Feinman and Brenden~M. Lake.
\newblock Learning inductive biases with simple neural networks.
\newblock {\em Proceedings of the 40th Annual Meeting of the Cognitive Science
  Society}, 2018.

\bibitem{finn2017model}
Chelsea Finn, Pieter Abbeel, and Sergey Levine.
\newblock Model-agnostic meta-learning for fast adaptation of deep networks.
\newblock In {\em International Conference on Machine Learning}, pages
  1126--1135. PMLR, 2017.

\bibitem{franceschi2018bilevel}
Luca Franceschi, Paolo Frasconi, Saverio Salzo, Riccardo Grazzi, and
  Massimiliano Pontil.
\newblock Bilevel programming for hyperparameter optimization and
  meta-learning.
\newblock In {\em International Conference on Machine Learning}, pages
  1568--1577. PMLR, 2018.

\bibitem{furlanello2018born}
Tommaso Furlanello, Zachary Lipton, Michael Tschannen, Laurent Itti, and Anima
  Anandkumar.
\newblock Born again neural networks.
\newblock In {\em International Conference on Machine Learning}, pages
  1607--1616. PMLR, 2018.

\bibitem{gidaris2019boosting}
Spyros Gidaris, Andrei Bursuc, Nikos Komodakis, Patrick P{\'e}rez, and Matthieu
  Cord.
\newblock Boosting few-shot visual learning with self-supervision.
\newblock In {\em International Conference on Computer Vision (ICCV)}, 2019.

\bibitem{gidaris2018unsupervised}
Spyros Gidaris, Praveer Singh, and Nikos Komodakis.
\newblock Unsupervised representation learning by predicting image rotations.
\newblock {\em arXiv preprint arXiv:1803.07728}, 2018.

\bibitem{guo2020attentive}
Yiluan Guo and Ngai-Man Cheung.
\newblock Attentive weights generation for few shot learning via information
  maximization.
\newblock In {\em IEEE Conference on Computer Vision and Pattern Recognition
  (CVPR)}, 2020.

\bibitem{guo2020broader}
Yunhui Guo, Noel~C Codella, Leonid Karlinsky, James~V Codella, John~R Smith,
  Kate Saenko, Tajana Rosing, and Rogerio Feris.
\newblock A broader study of cross-domain few-shot learning.
\newblock In {\em European Conference on Computer Vision}, pages 124--141.
  Springer, 2020.

\bibitem{helber2019eurosat}
Patrick Helber, Benjamin Bischke, Andreas Dengel, and Damian Borth.
\newblock Eurosat: A novel dataset and deep learning benchmark for land use and
  land cover classification.
\newblock {\em IEEE Journal of Selected Topics in Applied Earth Observations
  and Remote Sensing}, 12(7):2217--2226, 2019.

\bibitem{hinton2015distilling}
Geoffrey Hinton, Oriol Vinyals, and Jeff Dean.
\newblock Distilling the knowledge in a neural network.
\newblock {\em arXiv preprint arXiv:1503.02531}, 2015.

\bibitem{hou2019cross}
Ruibing Hou, Hong Chang, Bingpeng Ma, Shiguang Shan, and Xilin Chen.
\newblock Cross attention network for few-shot classification.
\newblock {\em arXiv preprint arXiv:1910.07677}, 2019.

\bibitem{mobilenet}
Andrew~G Howard, Menglong Zhu, Bo~Chen, Dmitry Kalenichenko, Weijun Wang,
  Tobias Weyand, Marco Andreetto, and Hartwig Adam.
\newblock Mobilenets: Efficient convolutional neural networks for mobile vision
  applications.
\newblock {\em arXiv preprint arXiv:1704.04861}, 2017.

\bibitem{hu2020empirical}
Shell~Xu Hu, Pablo~G Moreno, Yang Xiao, Xi~Shen, Guillaume Obozinski, Neil~D
  Lawrence, and Andreas Damianou.
\newblock Empirical bayes transductive meta-learning with synthetic gradients.
\newblock In {\em International Conference on Learning Representations (ICLR)},
  2020.

\bibitem{densenet}
Gao Huang, Zhuang Liu, Laurens Van Der~Maaten, and Kilian~Q Weinberger.
\newblock Densely connected convolutional networks.
\newblock In {\em Proceedings of the IEEE conference on computer vision and
  pattern recognition}, pages 4700--4708, 2017.

\bibitem{jamal2019task}
Muhammad~Abdullah Jamal and Guo-Jun Qi.
\newblock Task agnostic meta-learning for few-shot learning.
\newblock In {\em Proceedings of the IEEE/CVF Conference on Computer Vision and
  Pattern Recognition}, pages 11719--11727, 2019.

\bibitem{johnson2010mental}
Philip~N Johnson-Laird.
\newblock Mental models and human reasoning.
\newblock {\em Proceedings of the National Academy of Sciences},
  107(43):18243--18250, 2010.

\bibitem{kallenberg1997foundations}
Olav Kallenberg and Olav Kallenberg.
\newblock {\em Foundations of modern probability}, volume~2.
\newblock Springer, 1997.

\bibitem{kingma2014adam}
Diederik~P. Kingma and Jimmy Ba.
\newblock Adam: A method for stochastic optimization.
\newblock In {\em International Conference on Learning Representations (ICLR)},
  2014.

\bibitem{lee2019meta}
Kwonjoon Lee, Subhransu Maji, Avinash Ravichandran, and Stefano Soatto.
\newblock Meta-learning with differentiable convex optimization.
\newblock In {\em IEEE Conference on Computer Vision and Pattern Recognition
  (CVPR)}, 2019.

\bibitem{lee2018gradient}
Yoonho Lee and Seungjin Choi.
\newblock Gradient-based meta-learning with learned layerwise metric and
  subspace.
\newblock In {\em International Conference on Machine Learning}, pages
  2927--2936. PMLR, 2018.

\bibitem{li2017meta}
Zhenguo Li, Fengwei Zhou, Fei Chen, and Hang Li.
\newblock Meta-sgd: Learning to learn quickly for few-shot learning.
\newblock {\em arXiv preprint arXiv:1707.09835}, 2017.

\bibitem{lichtenstein2020tafssl}
Moshe Lichtenstein, Prasanna Sattigeri, Rogerio Feris, Raja Giryes, and Leonid
  Karlinsky.
\newblock Tafssl: Task-adaptive feature sub-space learning for few-shot
  classification.
\newblock In {\em European Conference on Computer Vision}, pages 522--539.
  Springer, 2020.

\bibitem{liu2020negative}
Bin Liu, Yue Cao, Yutong Lin, Qi~Li, Zheng Zhang, Mingsheng Long, and Han Hu.
\newblock Negative margin matters: Understanding margin in few-shot
  classification.
\newblock In {\em European Conference on Computer Vision (ECCV)}, 2020.

\bibitem{prototype}
Jinlu Liu, Liang Song, and Yongqiang Qin.
\newblock Prototype rectification for few-shot learning.
\newblock In {\em European Conference on Computer Vision (ECCV)}, 2020.

\bibitem{liu2018learning}
Yanbin Liu, Juho Lee, Minseop Park, Saehoon Kim, Eunho Yang, Sung~Ju Hwang, and
  Yi~Yang.
\newblock Learning to propagate labels: Transductive propagation network for
  few-shot learning.
\newblock {\em arXiv preprint arXiv:1805.10002}, 2018.

\bibitem{noroozi2016unsupervised}
Mehdi Noroozi and Paolo Favaro.
\newblock Unsupervised learning of visual representations by solving jigsaw
  puzzles.
\newblock In {\em European conference on computer vision}, pages 69--84.
  Springer, 2016.

\bibitem{phoo2021selftraining}
Cheng~Perng Phoo and Bharath Hariharan.
\newblock Self-training for few-shot transfer across extreme task differences.
\newblock In {\em International Conference on Learning Representations}, 2021.

\bibitem{qi2018low}
Hang Qi, Matthew Brown, and David~G Lowe.
\newblock Low-shot learning with imprinted weights.
\newblock In {\em Proceedings of the IEEE conference on computer vision and
  pattern recognition}, pages 5822--5830, 2018.

\bibitem{qiao2019transductive}
Limeng Qiao, Yemin Shi, Jia Li, Yaowei Wang, Tiejun Huang, and Yonghong Tian.
\newblock Transductive episodic-wise adaptive metric for few-shot learning.
\newblock In {\em Proceedings of the IEEE/CVF International Conference on
  Computer Vision}, pages 3603--3612, 2019.

\bibitem{qiao2018few}
Siyuan Qiao, Chenxi Liu, Wei Shen, and Alan~L Yuille.
\newblock Few-shot image recognition by predicting parameters from activations.
\newblock In {\em Proceedings of the IEEE Conference on Computer Vision and
  Pattern Recognition}, pages 7229--7238, 2018.

\bibitem{qin2019rethinking}
Zhenyue Qin, Dongwoo Kim, and Tom Gedeon.
\newblock Rethinking softmax with cross-entropy: Neural network classifier as
  mutual information estimator.
\newblock {\em arXiv preprint arXiv:1911.10688}, 2019.

\bibitem{tiered_imagenet}
Mengye Ren, Eleni Triantafillou, Sachin Ravi, Jake Snell, Kevin Swersky,
  Joshua~B Tenenbaum, Hugo Larochelle, and Richard~S Zemel.
\newblock Meta-learning for semi-supervised few-shot classification.
\newblock In {\em International Conference on Learning Representations (ICLR)},
  2018.

\bibitem{ritter2017cognitive}
Samuel Ritter, David~GT Barrett, Adam Santoro, and Matt~M Botvinick.
\newblock Cognitive psychology for deep neural networks: A shape bias case
  study.
\newblock In {\em International conference on machine learning}, pages
  2940--2949. PMLR, 2017.

\bibitem{imagenet}
Olga Russakovsky, Jia Deng, Hao Su, Jonathan Krause, Sanjeev Satheesh, Sean Ma,
  Zhiheng Huang, Andrej Karpathy, Aditya Khosla, Michael Bernstein,
  Alexander~C. Berg, and Li~Fei-Fei.
\newblock {ImageNet Large Scale Visual Recognition Challenge}.
\newblock In {\em International booktitle of Computer Vision (IJCV)}, 2015.

\bibitem{leo}
Andrei~A Rusu, Dushyant Rao, Jakub Sygnowski, Oriol Vinyals, Razvan Pascanu,
  Simon Osindero, and Raia Hadsell.
\newblock Meta-learning with latent embedding optimization.
\newblock In {\em International Conference on Learning Representations (ICLR)},
  2019.

\bibitem{prototypical_nets}
Jake Snell, Kevin Swersky, and Richard Zemel.
\newblock Prototypical networks for few-shot learning.
\newblock In {\em Advances in neural information processing systems (NeurIPS)},
  2017.

\bibitem{su2020does}
Jong-Chyi Su, Subhransu Maji, and Bharath Hariharan.
\newblock When does self-supervision improve few-shot learning?
\newblock In {\em European Conference on Computer Vision}, pages 645--666.
  Springer, 2020.

\bibitem{sung2018learning}
Flood Sung, Yongxin Yang, Li~Zhang, Tao Xiang, Philip~HS Torr, and Timothy~M
  Hospedales.
\newblock Learning to compare: Relation network for few-shot learning.
\newblock In {\em Proceedings of the IEEE conference on computer vision and
  pattern recognition}, pages 1199--1208, 2018.

\bibitem{tian2020rethinking}
Yonglong Tian, Yue Wang, Dilip Krishnan, Joshua~B. Tenenbaum, and Phillip
  Isola.
\newblock Rethinking few-shot image classification: a good embedding is all you
  need?
\newblock In {\em European Conference on Computer Vision (ECCV)}, 2020.

\bibitem{van2018inaturalist}
Grant Van~Horn, Oisin Mac~Aodha, Yang Song, Yin Cui, Chen Sun, Alex Shepard,
  Hartwig Adam, Pietro Perona, and Serge Belongie.
\newblock The inaturalist species classification and detection dataset.
\newblock In {\em Proceedings of the IEEE Conference on Computer Vision and
  Pattern Recognition}, pages 8769--8778, 2018.

\bibitem{verschueren2005everyday}
Niki Verschueren, Walter Schaeken, and Gery d’Ydewalle.
\newblock Everyday conditional reasoning: A working memory—dependent tradeoff
  between counterexample and likelihood use.
\newblock {\em Memory \& Cognition}, 33(1):107--119, 2005.

\bibitem{matching_net}
Oriol Vinyals, Charles Blundell, Timothy Lillicrap, Daan Wierstra, et~al.
\newblock Matching networks for one shot learning.
\newblock In {\em Advances in Neural Information Processing Systems (NeurIPS)},
  2016.

\bibitem{wang2020understanding}
Tongzhou Wang and Phillip Isola.
\newblock Understanding contrastive representation learning through alignment
  and uniformity on the hypersphere.
\newblock In {\em International Conference on Machine Learning}, pages
  9929--9939. PMLR, 2020.

\bibitem{simpleshot}
Yan Wang, Wei-Lun Chao, Kilian~Q Weinberger, and Laurens van~der Maaten.
\newblock Simpleshot: Revisiting nearest-neighbor classification for few-shot
  learning.
\newblock In {\em arXiv preprint arXiv:1911.04623}, 2019.

\bibitem{weinberger2006distance}
Kilian~Q Weinberger, John Blitzer, and Lawrence~K Saul.
\newblock Distance metric learning for large margin nearest neighbor
  classification.
\newblock In {\em Advances in neural information processing systems}, pages
  1473--1480, 2006.

\bibitem{meta_inat}
Davis Wertheimer and Bharath Hariharan.
\newblock Few-shot learning with localization in realistic settings.
\newblock In {\em Proceedings of the IEEE/CVF Conference on Computer Vision and
  Pattern Recognition}, pages 6558--6567, 2019.

\bibitem{ye2020few}
Han-Jia Ye, Hexiang Hu, De-Chuan Zhan, and Fei Sha.
\newblock Few-shot learning via embedding adaptation with set-to-set functions.
\newblock In {\em 2020 IEEE/CVF Conference on Computer Vision and Pattern
  Recognition (CVPR)}, pages 8805--8814. IEEE, 2020.

\bibitem{feat}
Han-Jia Ye, Hexiang Hu, De-Chuan Zhan, and Fei Sha.
\newblock Learning embedding adaptation for few-shot learning.
\newblock In {\em IEEE Conference on Computer Vision and Pattern Recognition
  (CVPR)}, 2020.

\bibitem{zagoruyko2016wide}
Sergey Zagoruyko and Nikos Komodakis.
\newblock Wide residual networks.
\newblock In {\em British Machine Vision Conference 2016}. British Machine
  Vision Association, 2016.

\bibitem{zhang2020deepemd}
Chi Zhang, Yujun Cai, Guosheng Lin, and Chunhua Shen.
\newblock Deepemd: Few-shot image classification with differentiable earth
  mover's distance and structured classifiers.
\newblock In {\em Proceedings of the IEEE/CVF Conference on Computer Vision and
  Pattern Recognition}, pages 12203--12213, 2020.

\bibitem{zhang2021iept}
Manli Zhang, Jianhong Zhang, Zhiwu Lu, Tao Xiang, Mingyu Ding, and Songfang
  Huang.
\newblock Iept: Instance-level and episode-level pretext tasks for few-shot
  learning.
\newblock In {\em Proc. Int. Conf. Learn. Represent.}, pages 1--16, 2021.

\bibitem{ziko2020laplacian}
Imtiaz Ziko, Jose Dolz, Eric Granger, and Ismail~Ben Ayed.
\newblock Laplacian regularized few-shot learning.
\newblock In {\em International Conference on Machine Learning}, pages
  11660--11670. PMLR, 2020.

\end{thebibliography}

\newpage
\appendix
\section{Baseline Implementation}
\label{appendix:baseline_details}
\myparagraph{Simple Baseline.} As stated above, the \emph{simple baseline} is constructed by training the feature extractor on base class with standard cross-entropy loss, which yield the objective function:
\vspace{-3pt}
\begin{equation}
{\color{blue}\mathcal{L}^{simple}} = \mathcal{L}_{clas}(\theta, \psi; D_b) = \frac{1}{N_b} \sum_{i=0}^{N_b}\sum_{j=0}^{|C_b|}y_{ij} \log p_\psi(j \vert f_\theta(\mathbf{x}_i))
\end{equation}
where $\theta$ and $\psi$ represent parameters of feature extractor and the linear classification, respectively.
\myparagraph{Rotation Baseline.} For SSL, we implement the \textit{"rotation baseline''} (\emph{Rot baseline}). In detail, during global classification training, we rotate each image with \textbf{all} predefined angles: $\mathbf{x}_i^r = T(\mathbf{x}_i, r) \forall r\in \mathcal{R}$  with $\mathcal{R} = \{0^\circ, 90^\circ, 180^\circ, 270^\circ\}$ and enforce the encoder to recognize the correct translation. Thus, the combined objective function of this baseline can be defined as:
\begin{equation}
{\color{red}\mathcal{L}^{rot}} = {\color{blue}\mathcal{L}^{simple}} + \lambda_{ssl} \mathcal{L}_{ssl}(\theta, \phi; D_b, \mathcal{R})
\end{equation}
Precisely, we utilize a seperate classifier $g_\phi$ on top of feature extractor $f_\theta$  to predict the pertubation of images via cross-entropy loss with four classes corresponding to four possible translations similar to \etal{Gidaris}~\cite{gidaris2019boosting}. Accordingly, the SSL loss function $\mathcal{L}_{ssl}$ is given by:
\begin{equation}
\mathcal{L}_{ssl}(\theta, \phi; D_b, \mathcal{R})  = \frac{1}{N_b\vert\mathcal{R}\vert} \sum_{i=0}^{N_b}\sum_{j\in\mathcal{R}}\sum_{k=0}^{\vert\mathcal{R}\vert} \mathbb{I}(k = j) \log p_\phi(k \vert f_\theta(\mathbf{x}_i^j))
\end{equation}
Here, $\mathbb{I}(\cdot)$ is the indicator function and $p_\phi(\cdot|f_\theta)$ denotes the (predicted) probability of rotation angle. 

\myparagraph{Rotation + KD Baseline.} The \textit{"rotatiton + kd baseline''} (\emph{Rot + KD baseline}) is constructed by combining knowledge distillation \cite{hinton2015distilling} with rotation classification. Explicitly, we first train the feature extractor following the configuration of \textit{rotation baseline}, and then exert the born-again strategy \cite{furlanello2018born} to perform knowledge distillation for $T$ generations. The loss function of this baseline is given by:
\begin{multline}
{\color{orange}\mathcal{L}^{rot+kd}} = {\color{red}\mathcal{L}^{rot}} + \lambda_{kd\_clas}\mathcal{L}_{kd\_clas}(\theta_t, \psi_t; D_b, \theta_{t-1}, \psi_{t-1}, \tau) + \\
\lambda_{kd\_ssl}\mathcal{L}_{kd\_ssl}(\theta_t, \phi_t; D_b, \mathcal{R}, \theta_{t-1}, \phi_{t-1}, \tau)
\end{multline}

The above objectives optimize $\mathcal{L}_{clas}$ and $\mathcal{L}_{ssl}$ with both groundtruth labels along with the prediction from teacher models \ie the model (of the same architecture) trained from the previous generation:
\begin{gather}
\mathcal{L}_{kd\_clas}(\theta_t, \psi_t; D_b, \theta_{t-1}, \psi_{t-1}, \tau) = \frac{1}{N_b} \sum_{i=0}^{N_b}\sum_{j=0}^{|C_b|}p_{\psi_{t-1}}(j\vert f_{\theta_{t-1}} (\mathbf{x}_i), \tau) \log p_{\psi_t}(j \vert f_{\theta_t}(\mathbf{x}_i), \tau)  \\
\mathcal{L}_{kd\_ssl}(\theta_t, \phi_t; D_b, \mathcal{R}, \theta_{t-1}, \phi_{t-1}, \tau)  = \frac{1}{N_b\vert\mathcal{R}\vert} \sum_{i=0}^{N_b}\sum_{j\in\mathcal{R}}\sum_{k=0}^{\vert\mathcal{R}\vert} p_{\phi_{t-1}}(k\vert f_{\theta_{t-1}} (\mathbf{x}_i^j), \tau) \log p_{\phi_t}(k \vert f_{\theta_t}(\mathbf{x}_i^j), \tau)
\end{gather}
Following \etal{Tian}~\cite{tian2020rethinking}, we distill the network for $T=2$ generations and use the \textbf{last} checkpoint of previous generation as the teacher for next generation. For simplicity, we set the hyperparameters $\lambda_{ssl} = \lambda_{kd\_ssl} = \lambda_{kd\_clas} = 1$ and $\tau = 4$ for \emph{all} experiments.

\section{On the importance of stop-gradient operator}
\label{sec:on_stop_gradient_operator}
In this section, we elaborate the bleak outcome when removing the stop-gradient operator of loss function in Equation \ref{eq:loss_poodle}. As discussed before, the soft-weight in \emph{``pull/push''} loss terms should be used to guide the degree of force. Directly optimize these soft weights would lead to different underlying objective that \textbf{jointly maximizing conditional entropy} as we show below.

First, we define the objective function (for each positive/negative term) in Equation \ref{eq:loss_poodle} as \emph{weighted distance} loss: 
\begin{equation}
\label{eq:loss_wd}
\mathcal{L}_{wd} = \sum_{k=1}^K \gamma \cdot d(\mathbf{w}_k, \mathbf{x}_i) \mathcal{S}_G[\,p(k\vert\mathbf{x}_i, \mathbf{W})]
\end{equation}
Furthermore we define the \emph{log-sum-exp} loss function as:
\begin{equation}
\label{eq:loss_lse}
\mathcal{L}_{lse} = 	- \log \sum_{k=1}^K \exp (-\gamma\cdot d(\mathbf{w}_k, \mathbf{x}_i))
\end{equation}
\begin{lemma} The two loss function $\mathcal{L}_{wd}$ and $\mathcal{L}_{lse}$ have same set of solutions.
\label{lemma:logsumexp}
\end{lemma}
\emph{Proof.} This can be proved by taking the gradient of two terms \wrt $j^\text{th}$ prototype. Particularly, the gradient of $\mathcal{L}_{lse}$ \wrt $\mathbf{w}_j$ is given by:
\begin{align}
\frac{\partial\mathcal{L}_{lse}}{\partial\mathbf{w}_j} &= - \frac{\partial\log \sum_{k=1}^K \exp (-\gamma\cdot d(\mathbf{w}_k, \mathbf{x}_i))}{\partial\mathbf{w}_j}\\
&= - \frac{1}{\sum_{k=1}^K\exp(-\gamma \cdot d(\mathbf{w}_k, \mathbf{x}_i))}\times \frac{\partial\sum_{k=1}^K\exp(-\gamma\cdot d(\mathbf{w}_k, \mathbf{x}_i))}{\partial\mathbf{w}_j} \\
&= - \frac{\exp(-\gamma\cdot d(\mathbf{w}_j, \mathbf{x}_i))}{\sum_{k=1}^K\exp(-\gamma \cdot d(\mathbf{w}_k, \mathbf{x}_i))} \times \frac{\partial(-\gamma\cdot d(\mathbf{w}_j, \mathbf{x}_i))}{\partial\mathbf{w}_j} \\
& = p(j\vert\mathbf{x}_i, \mathbf{W}) \cdot \frac{\partial \big(\gamma \cdot d(\mathbf{w}_j, \mathbf{x}_i)\big)}{\partial\mathbf{w}_j} \label{eq:loss_lse_implicit_cond_prob}
\end{align}
On the other hand, the gradient of $\mathcal{L}_{wd}$ term \wrt to $\mathbf{w}_j$ is given by:
\begin{align}
\frac{\partial\mathcal{L}_{wd}}{\partial\mathbf{w}_j} &= \frac{\partial \sum_{k=1}^K \gamma \cdot d(\mathbf{w}_k, \mathbf{x}_i) \mathcal{S}_G[\,p(k\vert\mathbf{x}_i, \mathbf{W})]}{\partial \mathbf{w}_j}\\
&= p(j\vert\mathbf{x}_i, \mathbf{W}) \frac{\partial\big(\gamma\cdot d(\mathbf{w}_j, \mathbf{x}_i)\big)}{\partial\mathbf{w}_j}
\end{align}
Thus, both loss function should optimize the same underlying objective. 
\vspace{10pt}
\begin{corollary}
	Taking out the \emph{stop-gradient} operator in ``weighted distance'' loss results in a objective that jointly minimizing conditional entropy of label given raw features and weighted distance (\ie the old loss function).
\end{corollary}
\emph{Proof:} We refer to this objective function as \emph{``Expected Distance''} loss. We can prove above corollary by rewriting the objective function as below:
\begin{align}
\mathcal{L}_{ed} &= \sum_{k=1}^K \gamma \cdot d(\mathbf{w}_k, \mathbf{x}_i) \,p(k\vert\mathbf{x}_i, \mathbf{W}) = - \sum_{k=1}^Kp(k\vert\mathbf{x}_i, \mathbf{W}) \log\Big[\exp\big(-\gamma\cdot d(\mathbf{w}_k, \mathbf{x}_i)\big)\Big]\\
&= -\sum_{k=1}^K \Big[p(k\vert\mathbf{x}_i, \mathbf{W})\log\frac{\exp(-\gamma\cdot d(\mathbf{w}_k,
\mathbf{x}_i))}{\sum_{k'}\exp(-\gamma\cdot d(\mathbf{w}_{k'}, \mathbf{x}_i))} + p(k\vert\mathbf{x}_i, \mathbf{W})\log\sum_{k'}\exp(-\gamma\cdot d(\mathbf{w}_{k'}, \mathbf{x}_i))\Big] \\
&= \underbrace{\Big(- \sum_{k=1}^K p(k\vert\mathbf{x}_i, \mathbf{W}) \log p(k\vert\mathbf{x}_i, \mathbf{W})\Big)}_\text{conditional entropy loss} - \underbrace{\log\sum_{k'}\exp\big(-\gamma\cdot d(\mathbf{w}_{k'}, \mathbf{x}_i)\big)}_\text{``old'' loss} \label{eq:combined_loss}
\end{align}
Using the Lemma \ref{lemma:logsumexp} and Equation \ref{eq:combined_loss}, we can observe that without the stop-gradient operator, underlying objective combining empirical (Monte-Carlo) estimate of conditional entropy of label given extracted features and ``weighted'' distance (\ie ``old'' loss). 

\vspace{10pt}
\myparagraph{Discussion.} As noted by \etal{Boudiaf} \cite{boudiaf2020transductive}, optimizing the conditional entropy loss requires special scrutiny, since the optima could result in trivial solutions on the simplex vertices \ie assigning all samples to a single class. Particularly, a small value of learning rate and fine-tuning the whole network (similar to \cite{dhillon2019baseline}) are crucial to prevent dramatical deterioration in performance. In their experiments \cite{boudiaf2020transductive}, they found that training the classifier with conditional entropy and cross-entropy loss signficant decrease the performance of few-shot learner. Beside, utilizing the conditional entropy for ``push'' loss does not affect the performance of classifier since it does not lead to collapsed solutions.

\section{Pseudo Code}
\label{appendix:pseudo_code}
We  provide the pseudo-code of \algo in a coding style similar  to Pytorch as in Algorithm \ref{alg:poodle_pytorch}. 
\begin{algorithm}[tbh]
	\caption{PyTorch-style pseudocode for \algo.}
	\label{alg:poodle_pytorch}
	
	\definecolor{codeblue}{rgb}{0.28,0.52,0.76}
	\definecolor{keywordorange}{rgb}{0.98, 0.51, 0.14}
	\lstset{
		basicstyle=\fontsize{9pt}{9pt}\ttfamily\bfseries,
		commentstyle=\fontsize{8pt}{8pt}\color{codeblue},
		keywordstyle=\bfseries\fontsize{12pt}{12pt}\color{keywordorange},
	}
	\begin{lstlisting}[language=python]
# f: classifier
# support: support images
# support_labels: labels of support images
# query: query images
# alpha: weight of positive term
# beta: weight of negative term

# sample negative data 
neg_data = sample_from_train_set()

# construct positive data
if transductive:
    n_support = support.size(0)
    pos_data = concatenate([support, query])
else:
    pos_data = support
        
for i in range(n_steps):
    # compute logits
    pos_out = f(pos_data)
    neg_out = f(neg_data)

    # compute POODLE loss
    pull_loss = sum(pos_out * softmax(pos_out).detach())
    push_loss = sum(neg_out * softmax(neg_out).detach())
    poodle_loss = alpha * pull_loss - beta * push_loss

    # compute CE loss
    ce_loss = sum(support_label * log_softmax(pos_out[: n_support]))
    loss = ce_loss + poodle_loss
	
    # optimization step
    loss.backward()
    optimizer.step()
	\end{lstlisting}
	
\end{algorithm}

\section{Additional results}
In this section, we provide more experimental results when applying \algo for various network architectures and conducting ablation study to verify its effectiveness under numerous configurations. \textbf{Note that by \algo, we refer to \algo-B \ie negative samples are drawn from base classes,  unless otherwise state.}
\subsection{Dataset} 
\label{appendix:dataset}
In this section, we briefly describe the datasets used for evaluating performance of \emph{cross-domain} FSL below.
\begin{itemize}[leftmargin=*]
    \item iNatural-2017 \cite{meta_inat}: This heavy-tailed dataset consists of $859,000$ images from over $5,000$ species of plants and animals. For \emph{disjoint} negative sampling, we follow the meta-iNat benchmark~\cite{meta_inat, simpleshot} \ie splitting the dataset to $908$ classes for sampling negative samples and $227$ classes for evaluation.
    \item EuroSAT \cite{helber2019eurosat}: This dataset covers total $27,000$ labeled images of Sentinel-2 satellite images, which consist of $10$ classes and the patches measure $64\times64$ pixels.
	\item ISIC-2018 (ISIC) \cite{codella2019skin}: This dataset covers dermoscopic images of skin lesions. Precisely, we use the training set for task $3$ (\ie lesion disease classification), which contains $10,015$ images with $7$ ground truth classification labels.
\end{itemize}
It is worth mentioning that, for the sake of simplicity, we apply the same image transformation as in \emph{mini}-Imagenet to these cross-domain datasets. Thus, the resolution of processed images of cross-domain datasets are $84\times84$ in our implementation. In contrast, the benchmark protocol using the image resolution of $224\times224$ \cite{guo2020broader}. A higher image resolution helps improve the performance of classifier significantly as we demonstrate in Section \ref{sec:higher_image_resolution}. Nevertheless, \algo successfully boosts the performance of all baselines in \emph{cross-domain} FSL by a large margin.

\subsection{Standard FSL}
\label{appendix:other_backbones}
In this section, we provide the experimental results of backbones other than Resnet-12. Specifically, we consider widely adopted architectures, namely Conv-4-512~\cite{matching_net}, WideResnet~\cite{zagoruyko2016wide}, Mobilenet~\cite{mobilenet}, and Densenet~\cite{densenet}. Particularly:
\begin{itemize}[leftmargin=*]
    \item \textbf{Conv-4-512}: We follow \cite{matching_net} to implement this architecture. More concretely, it is consisted of $4$ convolutional layers with hidden channels of $64$ and the output dimension is $512$.
    \item \textbf{WRN-28-10} \cite{zagoruyko2016wide}: We follow \cite{leo, simpleshot} and use the wide residual network with $28$ convolutional layers and widening factor of $10$.
    \item \textbf{DenseNet-121} \cite{densenet}: Similar to \cite{simpleshot}, we adopt the $121$-layers architecture while removing the first two-down sampling layers and using the kernel size of $3\times3$ for the first convolutional layer.
    \item \textbf{MobileNet} \cite{mobilenet}: We follow \cite{simpleshot} and use the standard MobileNet for ImageNet \cite{mobilenet} but remove the first two down-sampling layers of the network.
\end{itemize}
For all aforementioned architectures, we adopt the same training configuration of Resnet-12 as described in Section \ref{sec:experiments}, excepts for WRN-28-10 we use the batch size of $32$ for all datasets. Beside that, we use the negative samples from base classes \ie \algo-B in all experiments unless otherwise stated.

Table \ref{tab:various_network_architecture} shows the improvement of \algo with various backbones of different network architectures on \emph{mini}-Imagenet. We can see that our approach consistently enhance the accuracy of \textbf{all} baselines by a large margin \ie $2-4\%$ in $1$-shot protocol and $0.5-1\%$ in $5$-shot protocol. 

In Table \ref{tab:additional_benchmark_results_wrn}, we present the comparison between performance achieved by \textbf{WRN-28-10} with our approach and other algorithms on \emph{mini}- and \emph{tiered}-Imagenet. It can be observed that combining \algo with other techniques such as SSL and KD achieves comparable or outperform state-of-the-art techniques for both inductive/transductive inference on two datasets.

\begin{table}[h]
	\centering
    	\caption{Comparison to different baselines for the standard FSL (a.k.a in-domain FSL) on \textbf{\textit{mini}-ImageNet} with various network architectures. The results are evaluated with $10,000$ random sampled tasks. The $95\%$ confidence interval of $1$-shot and $5$-shot classification are roughly $0.2$ and $0.1$, respectively.}
	    \begin{tabular}{llp {1.5cm}p{1.5cm}p{1.5cm}p{1.5cm}p{1.5cm}}
		\toprule
		& & &\multicolumn{2}{c}{\textbf{Inductive}} & \multicolumn{2}{c}{\textbf{Transductive}} \\
		Network & Baseline & Variant & 1-shot & 5-shot & 1-shot & 5-shot\\
		\toprule
		\multirow{6}{*}{WRN-28-10} & \multirow{2}{*}{Simple} & CE Loss & 61.23 & 81.08 & 61.23 & 81.08 \\
		& & \algo & \cellcolor{pearDark!20}\textbf{64.94} & \cellcolor{pearDark!20}\textbf{81.88} & \cellcolor{pearDark!20}\textbf{71.49} & \cellcolor{pearDark!20}\textbf{83.48} \\
		\cmidrule{2-7}
		& \multirow{2}{*}{Rot}& CE Loss & 64.77 & 83.66 & 64.77 & 83.66 \\
		& & \algo & \cellcolor{pearDark!20}\textbf{68.27} & \cellcolor{pearDark!20}\textbf{84.45} & \cellcolor{pearDark!20}\textbf{75.24} & \cellcolor{pearDark!20}\textbf{85.95} \\
		\cmidrule{2-7}
		& \multirow{2}{*}{Rot + KD} & CE Loss & 67.12 & 84.15 & 67.12 & 84.15\\
		& & \algo & \cellcolor{pearDark!20}\textbf{69.67} & \cellcolor{pearDark!20}\textbf{84.84} & \cellcolor{pearDark!20}\textbf{77.30} & \cellcolor{pearDark!20}\textbf{86.34} \\
		\midrule
		
		\multirow{6}{*}{DenseNet-121} & \multirow{2}{*}{Simple} & CE Loss & 64.88 & 81.99 & 64.88 & 81.99 \\
		& & \algo & \cellcolor{pearDark!20}\textbf{66.53} &\cellcolor{pearDark!20}\textbf{82.55} & \cellcolor{pearDark!20}\textbf{75.55} & \cellcolor{pearDark!20}\textbf{84.62} \\
		\cmidrule{2-7}
		& \multirow{2}{*}{Rot}& CE Loss & 66.51 & 83.92 & 66.51 & 83.92 \\
		& & \algo & \cellcolor{pearDark!20}\textbf{68.68} & \cellcolor{pearDark!20}\textbf{84.52} & \cellcolor{pearDark!20}\textbf{77.85}
		& \cellcolor{pearDark!20}\textbf{86.49} \\
		\cmidrule{2-7}
		& \multirow{2}{*}{Rot + KD} & CE Loss & 68.66 & 84.17 & 68.66 & 84.17\\
		& & \algo & \cellcolor{pearDark!20}\textbf{70.29} & \cellcolor{pearDark!20}\textbf{84.80} & \cellcolor{pearDark!20}\textbf{78.57} & \cellcolor{pearDark!20}\textbf{84.49} \\
		\midrule
		
		\multirow{6}{*}{Conv-4-512} & \multirow{2}{*}{Simple} & CE Loss & 50.90 & 70.44 & 50.90 & 70.44 \\
		& & \algo & \cellcolor{pearDark!20}\textbf{53.60} & \cellcolor{pearDark!20}\textbf{71.08} & \cellcolor{pearDark!20}\textbf{58.61} & \cellcolor{pearDark!20}\textbf{72.78} \\
		\cmidrule{2-7}
		& \multirow{2}{*}{Rot}& CE Loss & 51.36 & 70.81 & 51.36 & 70.81 \\
		& & \algo & \cellcolor{pearDark!20}\textbf{54.04} & \cellcolor{pearDark!20}\textbf{71.69} & \cellcolor{pearDark!20}\textbf{58.84} & \cellcolor{pearDark!20}\textbf{73.24}  \\
		\cmidrule{2-7}
		& \multirow{2}{*}{Rot + KD} & CE Loss & 51.46 & 70.42 & 51.46 & 70.42\\
		& & \algo & \cellcolor{pearDark!20}\textbf{53.87} & \cellcolor{pearDark!20}\textbf{71.36} &  \cellcolor{pearDark!20}\textbf{58.50} & \cellcolor{pearDark!20}\textbf{72.87} \\
		\midrule
		
		\multirow{6}{*}{MobileNet} & \multirow{2}{*}{Simple} & CE Loss & 60.99 & 77.45 & 60.99 & 77.45 \\
		& & \algo & \cellcolor{pearDark!20}\textbf{61.75} & \cellcolor{pearDark!20}\textbf{77.87} & \cellcolor{pearDark!20}\textbf{70.63} & \cellcolor{pearDark!20}\textbf{80.04} \\
		\cmidrule{2-7}
		& \multirow{3}{*}{Rot}& CE Loss & 64.57 & 80.86 & 64.57 & 80.86  \\
		& & \algo & \cellcolor{pearDark!20}\textbf{65.45} & \cellcolor{pearDark!20}\textbf{81.37} & \cellcolor{pearDark!20}\textbf{74.81} & \cellcolor{pearDark!20}\textbf{83.50}\\
		\cmidrule{2-7}
		& \multirow{3}{*}{Rot + KD} & CE Loss & 65.41 & 80.60 & 65.41 & 80.60\\
		& & \algo-I & \cellcolor{pearDark!20}\textbf{66.02} & \cellcolor{pearDark!20}\textbf{81.18} & \cellcolor{pearDark!20}\textbf{74.55} & \cellcolor{pearDark!20}\textbf{83.09} \\
		\bottomrule
		
    	\end{tabular}
	\label{tab:various_network_architecture}
\end{table}

\begin{table}[h]
	\centering
	\caption{Comparison to the state-of-the-art methods on \textit{mini}-ImageNet, and \textit{tiered}-Imagenet using \textbf{inductive} and \textbf{transductive} settings on \textbf{WRN-28-10}. The results obtained by our models (\textcolor{pearDark}{pear}-shaded) are averaged over $10,000$ episodes.}
	\begin{tabular}{lccp{1.2cm}p{1.2cm}p{1.2cm}p{1.2cm}}
		\toprule
		& &\multicolumn{2}{c}{\textbf{\textit{mini}-ImageNet}} & \multicolumn{2}{c}{\textbf{\textit{tiered}-ImageNet}} \\
		Method & Transd. & 1-shot & 5-shot & 1-shot & 5-shot \\
		\toprule
		LEO \cite{leo}& \multirow{5}{*}{\xmark} & 61.8 & 77.6 & 66.3 & 81.4\\
        SimpleShot \cite{simpleshot}& & 63.5 & 80.3 & 69.8 & 85.3\\
        MatchNet \cite{matching_net}& & 64.0 & 76.3 & - & -\\
        CC+rot+unlabeled \cite{gidaris2019boosting} & & 64.0 & 80.7 & 70.5 & 85.0\\
        FEAT \cite{feat} & & 65.1 & 81.1 & 70.4 & 84.4 \\
        Simple & &  61.23 & 81.08 & 67.63 & 83.93 \\
        \rowcolor{pearDark!20} Simple + \algo & &  64.94 & 81.88 & 70.25 & 84.64 \\
		Rot + KD & &  67.12 & 84.15 & - & - \\
		\rowcolor{pearDark!20} Rot + KD + \algo& &  \textbf{69.67} & \textbf{84.84} & - & - \\
        \midrule
		AWGIM \cite{guo2020attentive}& \multirow{10}{*}{\cmark} & 63.1 & 78.4 & 67.7 & 82.8 \\
		Ent-min \cite{dhillon2019baseline} & & 65.7 & 78.4 & 73.3 & 85.5 \\
		SIB  \cite{hu2020empirical} & & 70.0 & 79.2 & - & - \\
		BD-CSPN \cite{prototype}& & 70.3 & 81.9 & 78.7 & 86.92 \\
		LaplacianShot  \cite{ziko2020laplacian}  & & 74.9 & 84.1 & 80.2 & 87.6 \\
		TIM-ADM \cite{boudiaf2020transductive} & & 77.5 & 87.2 & 82.0 & 89.7 \\
		TIM-GD \cite{boudiaf2020transductive} & & \textbf{77.8} & \textbf{87.4} & \textbf{82.1} & \textbf{89.8} \\
        \rowcolor{pearDark!20} Simple + \algo & &  71.49 & 83.48 & 76.27 & 85.83 \\
		\rowcolor{pearDark!20} Rot + KD + \algo& &  77.30 & 86.34 & - & - \\
    	\bottomrule
	\end{tabular}
	\label{tab:additional_benchmark_results_wrn}
\end{table}

\subsection{Results of reimplemented transductive algorithms}
\label{appendix:reproduce}
In this section, we report the accuracy of our reimplemented of transductive algorithms and their originally reported performance (balanced query set). In Table \ref{tab:compare_original_transductive}, we present the performance of our reproduced transductive algorithms (on Resnet-12). We can see that our implementation achieve comparable or higher accuracy than the original work even though we adopt the more efficient architecture (\ie Resnet-12 compared to Densenet).

\begin{table}[h]
	\centering
	\caption{Results of our implementations of various \textbf{transductive} methods compared to original works on \textbf{\textit{mini}-ImageNet}. The results of our implementations (\textcolor{pearDark}{pear}-shaded) are evaluated on \emph{Resnet-12} + Rot + KD baseline with 2,000 random sampled tasks.}
	\begin{tabular}{llcc}
	    \toprule
		Methods & Network & 1-shot & 5-shot \\
		\toprule
		\rowcolor{pearDark!20}Mean-shift & Resnet-12 & $73.49 \pm 0.55$ & $84.24 \pm 0.32$\\
		Mean-shift \cite{lichtenstein2020tafssl} & DenseNet-121 & $71.39 \pm 0.27$ & $82.67 \pm 0.15$\\
		\rowcolor{pearDark!20}Bayes k-means & Resnet-12 & $71.40 \pm 0.50$ & $83.79 \pm 0.31$\\
		Bayes k-means \cite{lichtenstein2020tafssl} & DenseNet-121 & $72.05 \pm 0.24$ & $80.34 \pm 0.17$\\
		\rowcolor{pearDark!20}Soft k-means & Resnet-12 & $74.55 \pm 0.49$ & $84.53 \pm 0.30$\\
		Soft k-means \cite{tiered_imagenet} & Conv-4-64 & $50.09 \pm 0.45$ & $64.59 \pm 0.28$\\
		\rowcolor{pearDark!20}CAN\_T & Resnet-12 & $71.04 \pm 0.53$ & $84.21 \pm 0.30$\\
		CAN\_T \cite{hou2019cross} & Resnet-12 & $67.19 \pm 0.55$ & $80.64 \pm 0.35$\\
		\rowcolor{pearDark!20}TIM & Resnet-12 & $77.69 \pm 0.55$ & $87.40 \pm 0.29$\\
		TIM \cite{boudiaf2020transductive} & Resnet-18 & $73.90 \pm n/a$ & $85.00 \pm n/a$\\
		\bottomrule
	\end{tabular}
	\label{tab:compare_original_transductive}
\end{table}

\subsection{Imbalanced query set}
So far, prior works in (transductive) FSL mainly concern with balanced query samples \ie numbers of query images for each class are equal. However, in practice, there is no guarantee that this setting is hold. In this section, we conduct experiments to benchmark the performance of different transductive algorithms under class-skew \ie imbalanced query data. We simulate the long-tail distribution of the query samples through the Dirichlet distribution. Precisely, for each novel task, we sample the number of queries for every category from a Dirichlet distribution with concentration $\kappa$ for all classes so that the total number of query images is always $75$ (similar to the experiments in the previous section).

In the ``pull'' loss term, we employ a \emph{``soft''} weights for minimizing distance between prototypes and samples without any explicit regularization for distance between prototypes. In conventional setup for transductive learning, the number of query samples  of each category are uniformly distributed, hence, the positive samples of each category implicitly constrains the prototype by ``pulling'' corresponding prototype.  Under an extremely imbalanced query set, the positive samples from the dominated class might pull all the prototypes to their region by learning similar features of samples from different classes. In that case, increasing the value of $\beta$ to prevent the collapsed solution is necessary.

Table \ref{tab:imbalanced} presents the results of different transductive learning algorithms under the imbalanced query set. The detailed performance of the reproduced algorithms compared to the original papers are reported in Section \ref{appendix:reproduce}. The coefficient of ``push'' term is set to $\alpha=1$. Since TIM \cite{boudiaf2020transductive} explicitly use the uniform distribution information of samples in the query set, its performance drastically drops in the imbalanced FSL. Other variants of K-means are more robust to the long-tail distribution, but they are far inferior to \algo with $\beta = 0.75$. 

\begin{table} 
	\centering
	\caption{Results of transductive learning on imbalance \textbf{query} data of \textit{mini}-ImageNet. The results obtained by our models (\textcolor{pearDark}{blue pearl}-shaded) are averaged over 2,000 episodes. $\kappa$ denotes the concentration parameter of the Dirichlet distribution.}
	\begin{tabular}{lcccccccc}
	    \toprule
		& \multicolumn{2}{c}{$\kappa=0.5$} & \multicolumn{2}{c}{$\kappa=1$} &\multicolumn{2}{c}{$\kappa=2$} & \multicolumn{2}{c}{$\kappa=5$}\\
		Methods & 1-shot & 5-shot & 1-shot & 5-shot & 1-shot & 5-shot. & 1-shot & 5-shot\\
		\toprule
		Inductive & 66.91 & 82.75 & 66.16 & 82.62 & 66.10 & 83.01 & 66.23 & 83.16 \\
		Mean-shift \cite{lichtenstein2020tafssl} & 68.88 & 80.54 & 70.31 & 81.82 & 71.78 & 83.16 &72.92 & 83.98 \\
		Bayes k-means \cite{lichtenstein2020tafssl} & 69.69 & 82.86 & 69.91 & 82.95 & 70.64 &83.54 & 71.39 & 83.83 \\
		Soft k-means \cite{tiered_imagenet} & 65.31 & 75.65 & 68.61 & 78.90 & 71.56 & 81.58 &73.45 & 83.33 \\
		CAN\_T \cite{hou2019cross} & 72.30 & 83.53 & 71.28 & 83.62 & 71.09 & 84.06 & 71.18 & 84.25\\
		TIM \cite{boudiaf2020transductive} & 47.65 & 52.79 & 54.86 & 61.32 & 61.84 & 68.73 & 68.74& 76.39 \\
	    \rowcolor{pearDark!20}\algo-B $\beta = 0.00$ & 61.16 & 76.40 & 66.02 & 79.88 & 70.03 &82.43 & 73.03 & 84.23 \\
	    \rowcolor{pearDark!20}\algo-B $\beta = 0.50$ & 64.56 & 81.01 & 69.14 & 83.63 & 72.91 &85.14 & \textbf{75.79} & 85.78 \\
		\rowcolor{pearDark!20}\algo-B $\beta = 0.75$ & 75.20 & 86.51 & \textbf{74.06} & 86.17 &\textbf{74.77} & \textbf{86.48} & 74.74 & \textbf{86.34} \\
		\rowcolor{pearDark!20}\algo-B $\beta = 1.00$ & \textbf{75.48} & \textbf{88.36} & 71.14 &\textbf{86.31} & 70.20 & 85.53 & 68.94 & 84.31 \\
		\bottomrule
	\end{tabular}
	\label{tab:imbalanced}
\end{table}

\subsection{Ablation study}
\subsubsection{Removing stop-gradient operator}
As discussed before in Section \ref{sec:on_stop_gradient_operator}, the stop-gradient operator is of paramount important to avoid trivial solutions. In this section, we consider different choice of usage of the stop-gradient operator namely not using it in both ``push/pull'' terms, use only with one of two term, and use it for both terms in Table \ref{tab:stop_gradient}.

From the table, we can see that the stop-gradient operator does not affect the performance much in inductive inference. It can be explained as the posterior distribution of support data already has a high probability for ``correct'' classes and the conditional entropy term does not affect much. In transductive inference, removing the stop-gradient operator on ``pull'' term results in noticeable decrease in accuracy as we articulate in Section \ref{sec:on_stop_gradient_operator}. On the other hand, taking out stop-gradient operator on ``push'' term does not worsen the performance since it does not lead to trivial solutions.

\begin{table}[h]
	\centering
	\caption{Results of different settings of stop-gradient in push and pull terms. The results are evaluated with $2,000$ random sampled tasks on \emph{Resnet-12} with Rot + KD baseline. The $95\%$ confidence interval of $1$-shot and $5$-shot classification are roughly $0.45$ and $0.3$ for \textbf{inductive} setting and $0.6$ and $0.4$ for \textbf{transductive}, respectively. Results of inductive baseline for $1$-shot and $5$-shot are $65.91 \pm 0.44$ and $82.95 \pm 0.30$, respectively.}
	\begin{tabular}{lcccp{1.2cm}p{1.2cm}p{1.2cm}p{1.2cm}}
		\toprule
		& \multicolumn{2}{c}{\textbf{Stop-grad}} &\multicolumn{2}{c}{\textbf{Inductive}} & \multicolumn{2}{c}{\textbf{Transductive}} \\
		Method & Pull & Push & 1-shot & 5-shot & 1-shot & 5-shot \\
		\toprule
		\multirow{4}{*}{\algo-B} & \cmark & \cmark & 67.52 & 83.71 & 77.58 & 85.87\\
		 & \xmark & \cmark & 67.50 & 83.56 & 62.04 & 82.87\\
    	 & \cmark & \xmark & 67.37 & 83.73 & 78.51 & 86.28\\
    	 & \xmark & \xmark & 67.37 & 83.64 & 72.04 & 85.08\\
    	\midrule
    	\multirow{4}{*}{\algo-R} & \cmark & \cmark & 67.80 & 83.50 & 77.25 & 85.52\\
		 & \xmark & \cmark & 67.80 & 83.39 & 66.55 & 83.65\\
    	 & \cmark & \xmark & 67.77 & 83.43 & 78.11 & 85.83\\
    	 & \xmark & \xmark & 67.77 & 83.36 & 73.29 & 84.77\\
		\bottomrule
	\end{tabular}
	\label{tab:stop_gradient}
\end{table}

\subsubsection{Sensitive to $\alpha$ and $\beta$}
\begin{figure}[h]
	\centering
	\begin{subfigure}[b]{0.49\textwidth}
		\centering
		\includegraphics[width=\linewidth]{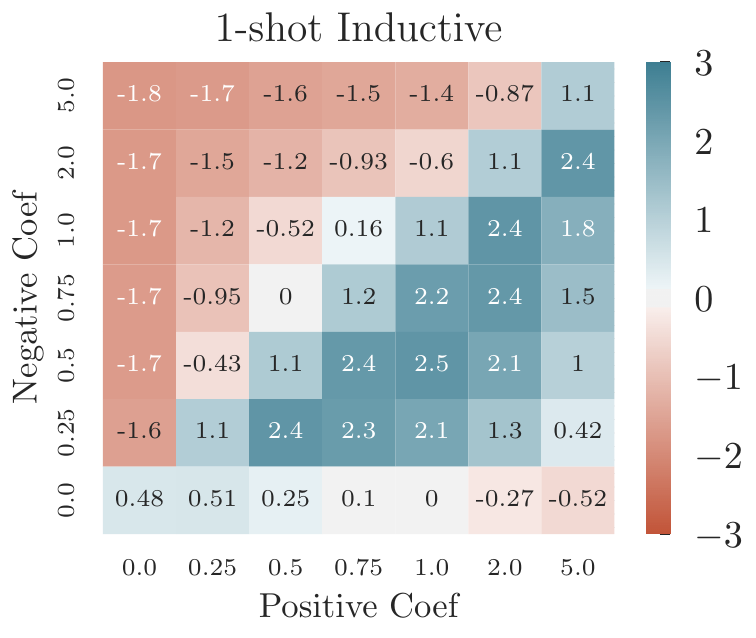}
		\caption{1-shot Inductive Inference}
		\label{fig:alpha_beta_ablation_a}
	\end{subfigure}
	\hfill
	\begin{subfigure}[b]{0.49\textwidth}
		\centering
		\includegraphics[width=\linewidth]{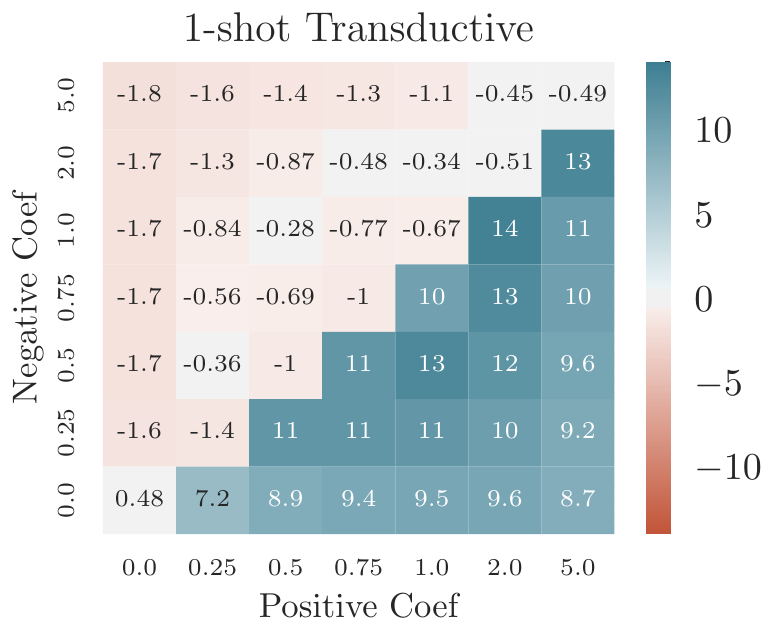}
		\caption{1-shot Transductive Inference}
		\label{fig:alpha_beta_ablation_b}
	\end{subfigure}
	\begin{subfigure}[b]{0.49\textwidth}
		\centering
		\includegraphics[width=\linewidth]{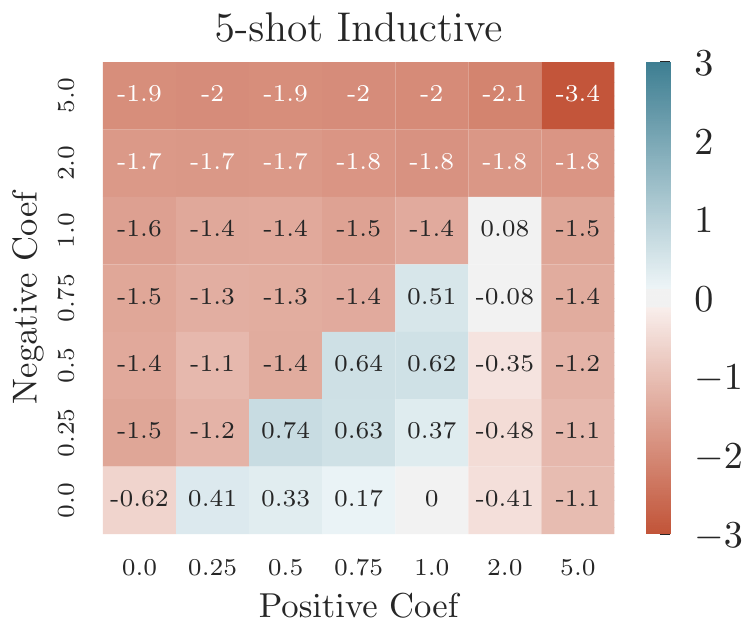}
		\caption{5-shot Inductive Inference}
		\label{fig:alpha_beta_ablation_c}
	\end{subfigure}
	\hfill
	\begin{subfigure}[b]{0.49\textwidth}
		\centering
		\includegraphics[width=\linewidth]{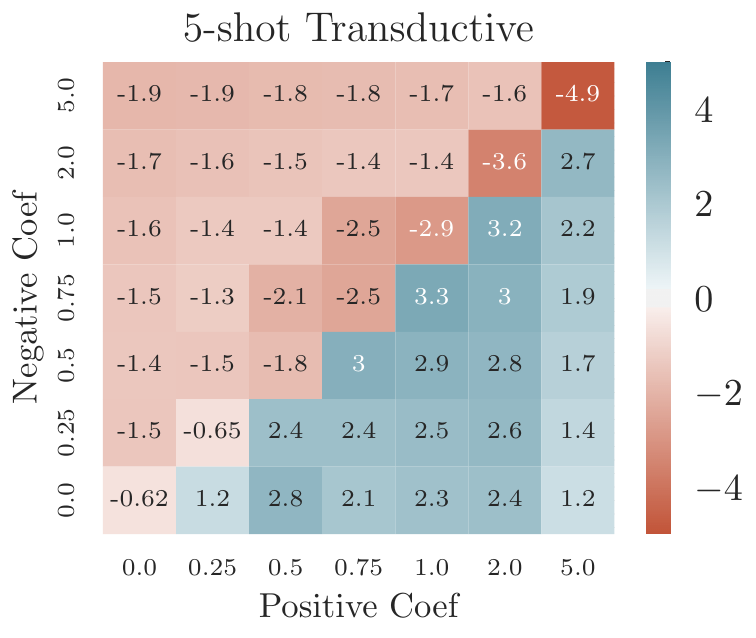}
		\caption{5-shot Transductive Inference}
		\label{fig:alpha_beta_ablation_d}
	\end{subfigure}
	\caption{Ablation study on the affect of positive and negative coefficient of \algo (\ie $\alpha$ and $\beta$) to \textbf{accuracy gain} of (simple baseline) \textbf{Resnet-12} on \emph{mini}-Imagenet compared to no fine-tuning at all with $10, 000$ random tasks. The baseline where we directly assign the prototype of each class to the mean of its support images obtains accuracy of $61.63 \pm 0.20$ and $80.78\pm0.11$ for $1$-shot and $5$-shot classification, respectively.}
	\label{fig:alpha_beta_ablation}
	\vspace{-5pt}
\end{figure}

So far, we only fixed the coefficient of ``push/pull'' loss with $\alpha=1$ and $\beta=0.5$, we now consider different configurations of these parameters to understand the sensitive of \algo to these hyper-parameters. Precisely, we report the accuracy gain of \algo with different combination of $\alpha$ and $\beta$ taken from the list $[0.0, 0.25, 0.5, 0.75, 1.0, 2.0, 5.0]$ as in Figure \ref{fig:alpha_beta_ablation}.

We can observe from Figure \ref{fig:alpha_beta_ablation} that \algo is robust to the change of $\alpha$ and $\beta$: the accuracy of network is improved as long as the value of $\alpha$ is larger than $\beta$. Furthermore, the improvement in accuracy usually does not change much when varying $\alpha$ and $\beta$. 

\subsubsection{Number of sampled negative data}
Intuitively, a larger number of negative samples will provide more informative cues to the classifer, however, they might require a notable memory footprint and have significant latency. We conduct experiment to quantify the impact of number of negative samples to the accuracy of \algo in Table \ref{tab:varying_num_neg_samples}. These results indicate that higher number of negative samples consistently leads to better performance, however, the gain in accuracy is relatively small.

\begin{table}[h]
	\centering
	\caption{Results of our methods with different number of out-of-distribution samples per task. The results are evaluated with $10,000$ random sampled tasks on \emph{Resnet-12} with Rot + KD baseline. The $95\%$ confidence interval of $1$-shot and $5$-shot classification are roughly $0.20$ and $0.10$ for both inductive and transductive inference. Results of inductive baseline for $1$-shot and $5$-shot are $66.32 \pm 0.20$ and $82.99 \pm 0.13$, respectively.}
	
	    \begin{tabular}{ccccc}
		\toprule
		 Number of & \multicolumn{2}{c}{\textbf{1-shot}} & \multicolumn{2}{c}{\textbf{5-shot}} \\
		negative samples & \algo-I & \algo-T & \algo-I & \algo-T\\
		\toprule
		50 & 67.61 & 77.38 & 83.54 & 85.56 \\
		100 & 67.71 & 77.61 & 83.62 & 85.64 \\
		200 & 67.77 & 77.69 & 83.69 & 85.70 \\
		500 & \textbf{67.84} & \textbf{77.73} & \textbf{83.71} & \textbf{85.73} \\
		\bottomrule
		
    	\end{tabular}
	\label{tab:varying_num_neg_samples}
\end{table}

\subsubsection{Higher image resolution}
\label{sec:higher_image_resolution}
Existed works in FSL usually employ the standard image pre-processing schemes such as resizing to $84\times84$ pixels. We consider using a slightly different setup where we resize the image to a higher resolution than $84$ during training and testing. We keep all other hyper-parameters and configurations as in \emph{simple baseline}. We report the accuracy of these models with \algo in both inductive and transductive inference in Table \ref{tab:varying_img_resolution}. It can be observed that the higher resolution usually leads to better performance. Nevertheless, \algo successfully raises the accuracy in all cases up to $5\%$ in inductive 1-shot classification.

\begin{table}[h]
	\centering
    	\caption{Comparing the performance of simple baseline on \emph{mini}-Imagenet with different image resolutions. \algo-I and \algo-T indicate the result of our algorithm in \textbf{inductive} and \textbf{transductive} inference. The results are evaluated with $10,000$ random sampled tasks. The $95\%$ confidence interval of $1$-shot and $5$-shot classification are roughly $0.2$ and $0.1$, respectively. \textbf{Bold} numbers indicate significant improvement (\ie p-value $\le 0.05$) compared to \emph{weakest} corresponding entries.}
    	\resizebox{\textwidth}{!}{
	    \begin{tabular}{cccccccc}
		\toprule
		& & \multicolumn{3}{c}{\textbf{1-shot}} & \multicolumn{3}{c}{\textbf{5-shot}} \\
		Network & Resolution & CE Loss & \algo-I & \algo-T & CE Loss & \algo-I & \algo-T\\
		\toprule
		\multirow{4}{*}{Resnet-12} & $84\times84$ & 61.63 & 64.08 & 74.46 & 80.78 & 81.41 & 83.70 \\
		& $140\times140$ & 62.66 & 65.90 & 75.08 & 82.06 & 82.89 & 84.88 \\
		& $180\times180$ & \textbf{64.85} & 67.05 & \textbf{76.17} & 82.52 & 83.19 & \textbf{85.31} \\
		& $224\times224$ & 62.90 & \textbf{67.22} & 74.46 & \textbf{82.59} & \textbf{83.61} & 85.23 \\

		\midrule
		\multirow{3}{*}{WRN-28-10} & $84\times84$ & 61.23 & 64.94 & \textbf{71.49} & 81.08 & 81.88 & 83.48 \\
		& $140\times140$ & \textbf{61.76} & 66.27 & 71.23 & \textbf{81.73} & \textbf{82.70} & \textbf{84.05} \\
		& $180\times180$ & 61.72 & \textbf{66.34} & 70.61 & 81.67 & 82.65 & 83.90 \\
		\bottomrule
		
    	\end{tabular}}
	\label{tab:varying_img_resolution}
\end{table}

%


\subsubsection{Different number of finetuning steps}
In previous experiments, we fixed the number of fine-tuning steps when employing \algo with $T=250$ steps. We now consider different numbers of fine-tuning steps and evaluate its impact on the final performance of the classifier. Particularly, we report the accuracy of the classifier fine-tuned with \algo and standard cross-entropy loss with various number of steps in Table \ref{tab:varying_finetuning_steps}. In our experiments, we find that \algo is not sensitive to number of gradient updates and do not requires a high number of steps to achieves good performance. This finding allows us to accelerate the fine-tuning steps and reduce the computational cost for inference phase.
\begin{table} 
	\centering
	\caption{Results of our methods with different number of update iterations. The results are evaluated with $2,000$ random sampled tasks on \emph{Resnet-12} with Rot + KD baseline. The $95\%$ confidence interval of $1$-shot and $5$-shot classification are roughly $0.45$ and $0.3$ for \textbf{inductive} setting and $0.5$ and $0.3$ for \textbf{transductive}, respectively. Results of inductive baseline for $1$-shot and $5$-shot are $65.91 \pm 0.44$ and $82.95 \pm 0.30$, respectively. \textbf{Bold} numbers indicate significant improvement (\ie p-value $\le 0.05$) compared to \emph{weakest} corresponding entries.}
	\begin{tabular}{ccccccc}
		\toprule
		& \multicolumn{3}{c}{\textbf{1-shot}} & \multicolumn{3}{c}{\textbf{5-shot}} \\
	    Finetuning steps & CE Loss & \algo-I & \algo-T & CE Loss & \algo-I & \algo-T\\
		\toprule
		50 & 66.18 & 67.61 & 74.75 & \textbf{83.05} & 83.72 & 85.57 \\
		100 & 66.17 & 67.54 & 77.03 & 82.94 & 83.71 & 85.78 \\
		250 & 66.17 & 67.52 & 77.58 & 82.76 & 83.69 & 85.87 \\
		400 & 66.17 & 67.49 & \textbf{77.63} & 82.68 & 83.70 & 85.87 \\
		\bottomrule
	\end{tabular}
	\label{tab:varying_finetuning_steps}
\end{table}

\end{document}